\documentclass[table]{article}

\usepackage[final,nonatbib]{neurips_2024}
\usepackage[numbers,sort&compress]{natbib}
\PassOptionsToPackage{hyphens}{url}

\usepackage[utf8]{inputenc}
\usepackage[T1]{fontenc}
\usepackage{xcolor}
\usepackage{hyperref}
\hypersetup{hidelinks}
\usepackage{graphicx}
\usepackage{caption}
\usepackage{booktabs}
\usepackage{colortbl}
\usepackage{amsmath}
\usepackage{amssymb}
\usepackage{amsthm}
\usepackage{pifont}
\usepackage{algorithm}
\usepackage{algorithmic}
\usepackage{float}
\usepackage{enumitem}
\usepackage{listings}
\usepackage[most]{tcolorbox}
\usepackage{placeins}
\usepackage{microtype}
\usepackage{wrapfig}
\usepackage{ragged2e}

\newtheorem{proposition}{Proposition}
\newtheorem{theorem}{Theorem}
\newtheorem{lemma}{Lemma}

\theoremstyle{definition}
\newtheorem{definition}{Definition}

\theoremstyle{remark}
\newtheorem{remark}{Remark}

\definecolor{vistablue}{HTML}{2E5AA8}
\definecolor{visiblegreen}{HTML}{2E7D32}
\definecolor{archiveblue}{HTML}{1E5AA8}
\newcommand{\statevisible}{\textcolor{visiblegreen}{\textsc{visible}}}
\newcommand{\statearchived}{\textcolor{archiveblue}{\textsc{archived}}}
\newcommand{\methodhead}[1]{\noindent{\textbf{#1}}}
\tcbset{
  promptbox/.style={
    colback=vistablue!4,
    colframe=vistablue!55,
    boxrule=0.5pt,
    arc=1.5mm,
    outer arc=1.5mm,
    left=2mm, right=2mm, top=1mm, bottom=1mm,
    fonttitle=\bfseries\small,
    before skip=5pt, after skip=5pt,
  }
}
\lstdefinestyle{prompt}{
  basicstyle=\ttfamily\fontsize{7.6pt}{8.4pt}\selectfont,
  frame=none,
  breaklines=true,
  breakatwhitespace=true,
  breakindent=0pt,
  columns=fullflexible,
  keepspaces=true,
  numbers=none,
  xleftmargin=2pt,
}

\newcommand{\yes}{\ding{51}}
\newcommand{\no}{\ding{55}}
\newcommand{\pmark}{$\sim$}

\newcommand{\method}{VISTA}
\newcommand{\sms}{VISTA}
\newcommand{\loca}{LOCA-Bench}
\newcommand{\ama}{AMA-Bench}

\title{LLM Agents Are Latent Context Managers:\protect\\
Eliciting Self-Managed Context via State Proprioception}

\author{
  \textbf{Binyan Xu$^{1,2,*}$},
  \textbf{Haitao Li$^{2,\dagger}$},
  \textbf{Kehuan Zhang$^{1}$} \\
  $^{1}$The Chinese University of Hong Kong,
  $^{2}$LIGHTSPEED \\
  \texttt{\{binyxu, khzhang\}@ie.cuhk.edu.hk},
  \texttt{729156675@qq.com} \\
  $^{*}$Work done during an internship at Tencent.
  $^{\dagger}$Corresponding author.
}

\begin{document}
\maketitle

\begin{abstract}
Long-horizon tool agents are bottlenecked by how their context grows toward the
limits of the context window. Recent systems make context management agent- or
system-controlled, but they either learn compression policies that discard
evidence or manage context in a layer the agent never sees. We argue that both
miss a more basic gap: frontier language models are proprioceptively blind to
their own context. From the prompt alone they cannot reliably infer block size,
recency, or the remaining budget, all of which are needed for keep-or-archive decisions. We introduce
\method{} (Visible Internal State for Tool Agents), a training-free,
model-agnostic layer that represents working memory as typed addressable blocks,
surfaces a runtime dashboard of token usage, recency, archive status, and remaining budget, and
archives blocks as recoverable full-fidelity payloads. On \loca{},
BrowseComp-Plus, and GAIA, the same untrained interface transfers across
1M-, 100K-, and 10K-scale trajectories. On \loca{} it lifts
Gemini-3-Flash from 22.7 to 50.7\%, reaches 58.0\% on BrowseComp-Plus, and
remains competitive on GAIA. Gains grow with context pressure and transfer
across backbones, while ablations confirm that the dashboard matters beyond
archive and recovery tools.
\end{abstract}

\begin{figure}[H]
\centering
\includegraphics[width=1.0\textwidth]{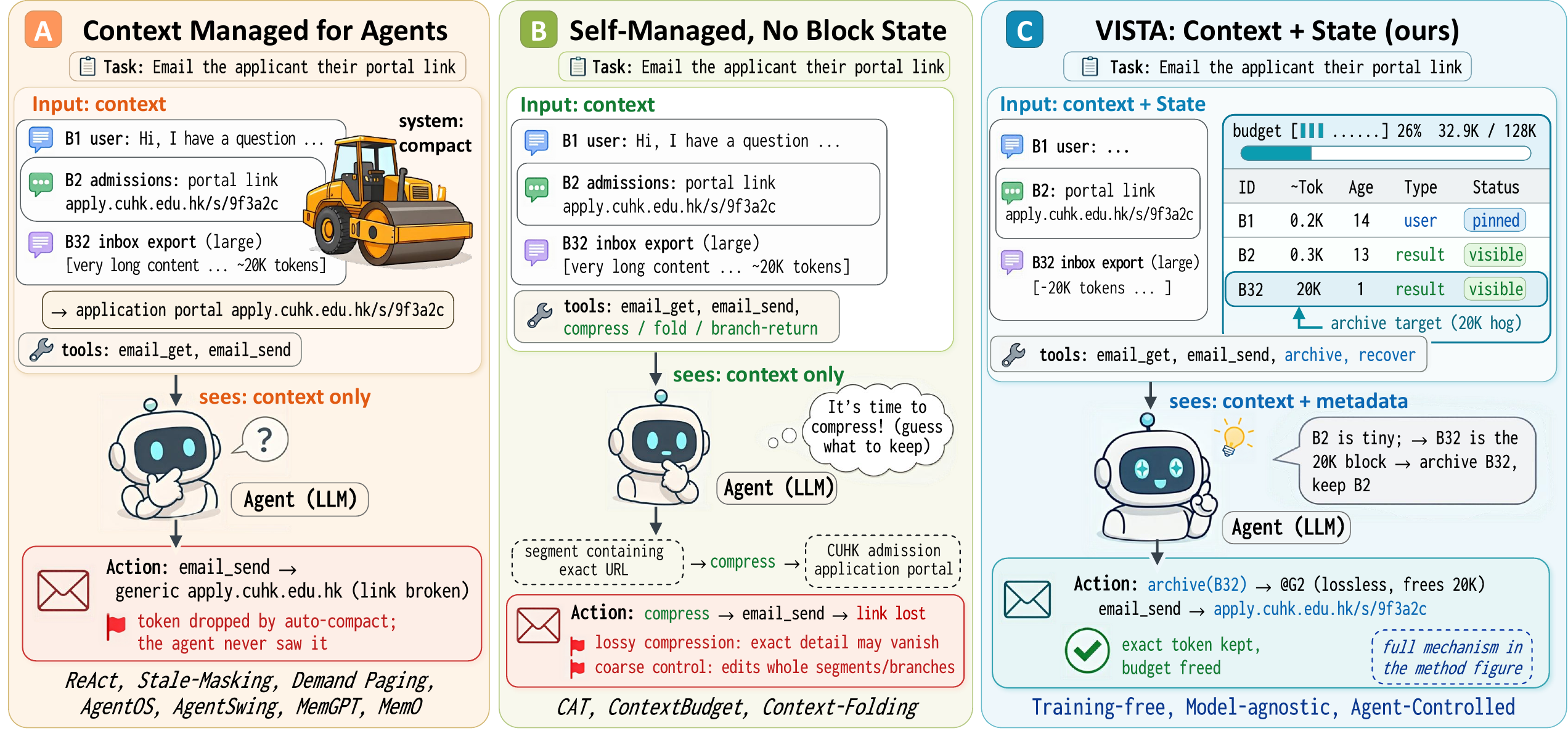}
\caption{\textbf{Who manages context, and on what information.} Fixed rules
compact context the agent cannot see, and blind self-management guesses without
state. \method{} surfaces per-block metadata, so the agent archives the large
block losslessly.}
\label{fig:teaser}
\end{figure}

\section{Introduction}

Language agents operate over stateful tasks such as filling spreadsheets from
web and email evidence, modifying databases, preparing application materials,
debugging code, and coordinating business workflows~\citep{locabench2026,contextastool2025,xu2026agent}.
Their context is working memory. It accumulates tool evidence, stale
observations, failed attempts, user constraints, hypotheses, file paths, and
action contracts that must remain correct many steps later
\citep{packer2023memgpt,activecontext2026,contextbudget2026}. As the task runs,
working memory grows until it crowds or overflows the context window, a pressure
also studied in long reasoning systems that summarize or carry state across
computation~\citep{memento2026,reasoningcache2026,markovianthinker2025}. The
agent must decide what to keep visible, what to set aside, and what to recover.
How this growing context is managed determines whether long-horizon agents
succeed.

Existing approaches differ in who makes these decisions. One family keeps the
decision outside the agent. Stale-observation masking hides old tool outputs by
rule~\citep{staleobs2026}, and OS-style layers page or evict context beneath the
agent. These layers track statistics such as size, age, and usage, but only
inside the runtime. The agent cannot inspect them, and a fixed rule cannot know
which evidence will matter later. A second family moves the decision into the
agent and learns it from data. Context-as-a-tool fine-tunes a compressor, and
budget-aware methods train compression policies with reinforcement
learning~\citep{contextastool2025,activecontext2026,contextbudget2026,contextfolding2025}.
These methods can improve performance, but they often discard evidence through
summarization or deletion and are tied to the training setting. Across both
families, the agent can read block contents but remains context-state blind: the
prompt omits the runtime state needed for a keep-or-archive decision, including block size,
recency, archive status, and remaining budget. Figure~\ref{fig:teaser} contrasts these families with our
approach on one task.

\begin{wrapfigure}[18]{r}{0.49\textwidth}
\centering
\captionsetup{font=small}
\includegraphics[width=0.48\textwidth]{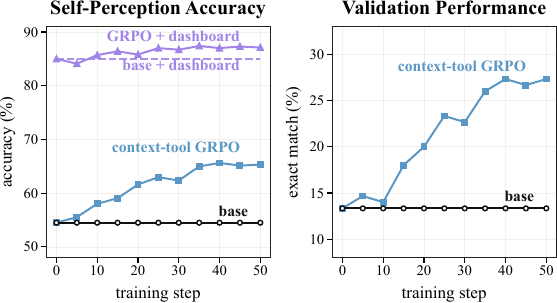}
\caption{\textbf{RL implicitly learns self-management.} Across training, task
performance and dashboard-free self-perception improve together, while
dashboard-aided perception remains high. This gap suggests that the missing
ingredient is observable state, motivating us to expose it explicitly instead of
asking the policy to learn it implicitly.}
\label{fig:rl-self-perception}
\end{wrapfigure}
We take an elicitation view. We hypothesize that capable models already contain
context-management competence from pretraining on note-taking, retrieval, and
reorganization traces, and that a missing interface, not a missing policy, is
the bottleneck. Context management is a meta-tool decision over the
agent's own working memory, made under partial observability. The agent must
choose what to keep or externalize while the prompt omits the runtime state that
governs the choice. Learned policies or distilled agent skills can compensate
through training~\citep{xu2026multi}, but this entangles what information should
be exposed with what policy should act on it.
Figure~\ref{fig:rl-self-perception} makes this motivation concrete: RL-based
self-management implicitly improves both task performance and the model's
ability to recognize when its context needs management, whereas supplying the
state dashboard makes that judgment easy even before adaptation. We therefore
make the state explicit and express when-to-manage criteria as a rubric-style
instruction. \method{} elicits these latent capabilities without requiring
training or model-specific adaptation, while remaining compatible with RL.

This proprioceptive view implies three requirements. The interface must expose per-block token
cost, recency, archive status, and remaining budget. It must be reversible,
because one-way deletion or summarization can remove evidence needed later. It
must be model-agnostic, so gains reflect elicitation through the interface,
not training for one backbone or domain.

We introduce \method{} (Visible Internal State for Tool Agents), a context layer
that represents working memory as typed, addressable blocks and surfaces a
dashboard with per-block token usage, recency, archive status, and budget. The
dashboard is a proprioceptive view of the agent's context state. The
agent can archive bulky blocks as external payloads with stable handles and
recover exact bytes on demand. Archived payloads are exact transcripts, so
removing a block from the prompt does not destroy it. We prove that both underlying
resources are necessary: recoverability, because discarded evidence cannot
otherwise be restored; and sufficiently informative proprioceptive state,
because a size-blind manager over-archives or under-recovers even given recovery
(Proposition~\ref{prop:recover} and Theorem~\ref{thm:sep}). VISTA realizes both
resources in one agent-facing interface.
\method{} requires no training and wraps any backbone.

Empirically, Table~\ref{tab:main-results} tests \method{} across million-,
100K-, and 10K-token trajectories in \loca{}, BrowseComp-Plus, and GAIA. In
\loca{}~\citep{locabench2026}, it solves 38/75 tasks versus 17 for ReAct and 32
for Claude Code, with lower trajectory cost than Claude Code. On
BrowseComp-Plus, it reaches 58.0\% versus 52.0\% for the strongest baseline, and
remains competitive on GAIA. Pressure sweeps show that the advantage grows on
long trajectories, where \method{} cuts active-context overhead while improving
accuracy. The same untrained layer improves all four tested backbones, and
ablations show that the dashboard matters beyond archive and recovery tools. In
transfer, \method{} also reaches higher F1 than the specialized \ama{} agent
without memory tuning.

This paper makes three contributions.
\begin{itemize}
    \item We frame context management as a meta-tool decision under partial
    observability and identify context proprioception as the missing interface:
    LLM agents are context-state blind because their prompts omit reliable runtime state.
    \item We introduce \method{}, a training-free context layer
    whose dashboard exposes per-block metadata and pairs it with lossless
    archive and recovery. Two matched separations prove that recoverability and
    sufficiently informative proprioceptive state are jointly necessary; VISTA
    supplies both in a single agent-facing interface.
    \item We show across \loca{}, BrowseComp-Plus, and GAIA that \method{}
    elicits self-management across trajectory scales and backbones, isolate the
    dashboard with ablations, and demonstrate transfer on \ama{}.
\end{itemize}

\section{Methodology}
\label{sec:methodology}

\begin{figure*}[t]
\centering
\includegraphics[width=1.0\textwidth]{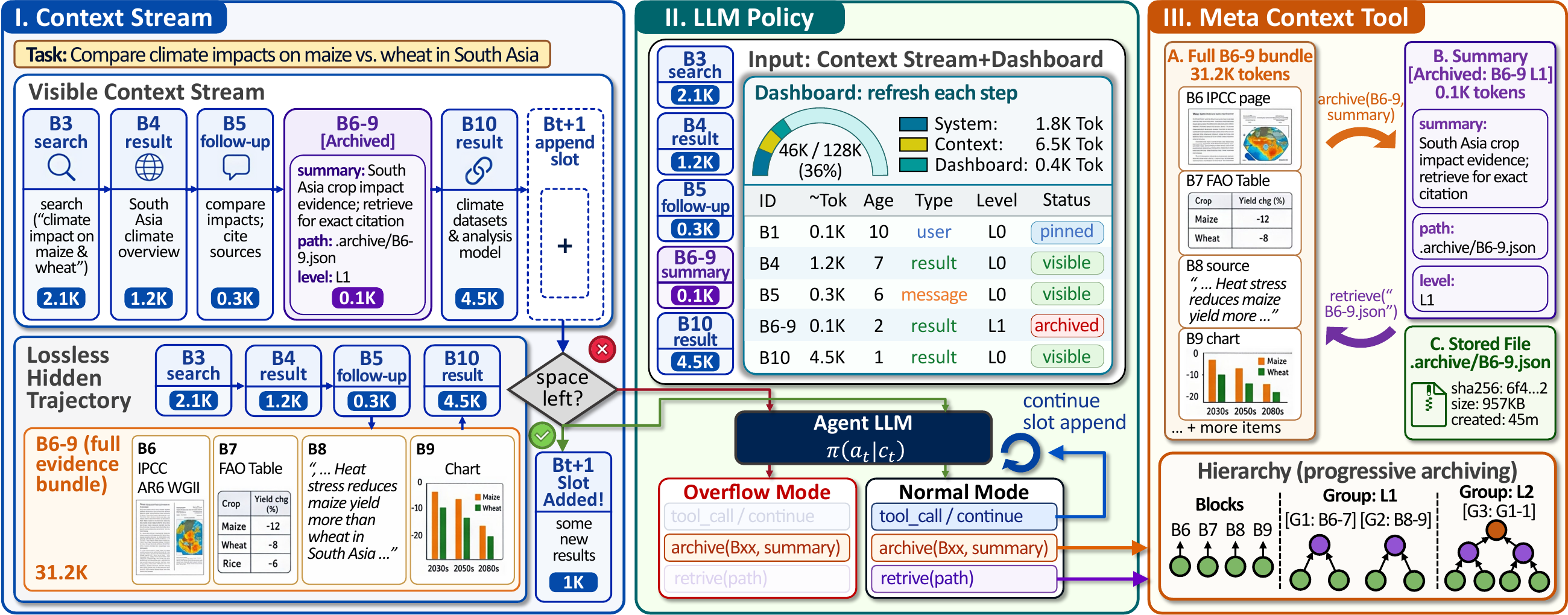}
\caption{\textbf{\method{} architecture.} Messages and tool outputs become
addressable blocks. The dashboard exposes budget and handles to the agent, while
archived payloads remain recoverable outside the active prompt.}
\label{fig:system}
\end{figure*}

\method{} treats context management as a meta-tool decision over the agent's own
working memory. Figure~\ref{fig:system} shows the three-stage loop: a context
stream, a refreshed dashboard, and archive/recovery tools. The goal is to make
context state perceptible to an unmodified model without fine-tuning,
model-specific changes, or destroyed evidence.

\subsection{Problem Setup}

At step $t$, a tool agent has a task goal $g$, raw interaction history $H_t$,
environment tools $\mathcal{T}_{\mathrm{env}}$, and a context budget $B$. We
write the history as action-observation pairs
$H_t=(a_1,o_1,\ldots,a_{t-1},o_{t-1})$. A standard ReAct-style harness
serializes this append-only history into the next model input. Once the
serialized history exceeds $B$, the harness must truncate, clear, mask, or
summarize prior content. These interventions conflate what remains visible,
what is preserved exactly, and what can be recovered later.

\method{} separates these choices through a workspace
$W_t=(V_t,A_t)$ of visible blocks and recoverable archived payloads. The model
does not act on the raw transcript directly; it acts on the
workspace rendering
\begin{equation}
\label{eq:workspace-render}
  \begin{aligned}
  \widetilde C_t &= \operatorname{render}(V_t)
      \oplus \operatorname{handles}(A_t)
      \oplus D_t,\\
  C_t &= \operatorname{preflight}(\widetilde C_t,B),\qquad
  a_t \sim \pi_\theta(\cdot \mid C_t),\qquad |C_t|\le B .
  \end{aligned}
\end{equation}
Here $D_t$ is the dashboard: a factual ledger of block IDs, token estimates,
recency, block type, archive level, status, and remaining budget.
The candidate $\widetilde C_t$ may exceed the budget. In that case,
$\operatorname{preflight}$ emits a budget-safe management rendering $C_t$ that
retains the dashboard, actionable handles, and compact stubs while disabling
ordinary task tools; otherwise it returns the ordinary rendering.
The raw transcript is still logged for evaluation, but the agent's working
memory is the workspace, not the append-only history.

\methodhead{Workspace invariants.}
The harness enforces a budget constraint, $|C_t|\le B$, after preflight and
final assembly; every actionable unit must have a stable block ID or handle; and
archived payload bytes remain recoverable unless the agent explicitly deletes
them. The dashboard reports the same token estimates used by these checks, so
the agent sees the state that the harness will enforce.

\subsection{Context Stream}

The left panel of Figure~\ref{fig:system} shows how \method{} rewrites the
transcript as a block stream. The agent-facing state has two storage regimes:
\begin{enumerate}[label=(\roman*),leftmargin=18pt,itemsep=1pt,topsep=2pt]
\item \statevisible{} blocks appear in the active prompt with exact content;
pinned blocks are a protocol-required subset of this state.
\item \statearchived{} blocks are replaced by compact handles and summaries,
while their original bytes are stored externally.
\end{enumerate}
Deletion is an explicit agent action, not a storage regime: it removes
the selected content permanently. A separate hard-budget guard may replace
already-visible raw tool results with recoverable offloaded placeholders; a new
tool result that still cannot fit is rejected with a compact notice and can be
rerun after context cleanup or with a narrower query. These safeguards are
distinct from the agent's archive decision.

This stream gives the agent an address space for context decisions. When a
bundle of evidence becomes too large, the stream keeps a compact handle while
the exact payload remains in the hidden trajectory. New observations are
admitted only when the resulting request remains budget-safe.

\methodhead{Structure preservation.}
The visible stream is the next-call working set, while the hidden trajectory
stores exact payloads that may be needed later. Tool results remain linked to their
assistant tool calls, and archived results are rendered in protocol-valid form.
If a parent call is also archived, the placeholder becomes ordinary context
instead of an orphaned tool response.

\subsection{LLM Policy}

\methodhead{Dashboard input.}
The middle panel shows what the model receives: the visible context stream plus
a dashboard. The dashboard is regenerated after every tool result is registered,
so the agent acts on the current workspace state. It is a ledger over blocks,
not a memory oracle: it exposes runtime state created by the harness or
by the agent's earlier actions and does not add hidden task evidence.

\methodhead{Unified action space.}
The same model policy chooses ordinary environment actions and context actions:
\begin{equation}
\label{eq:action-space}
\begin{aligned}
\mathcal{T}_{\mathrm{ctx}}
&=\{\operatorname{archive}(\mathcal{S},\rho),\
    \operatorname{delete}(\mathcal{S})\},\\
a_t &\in \mathcal{T}_{\mathrm{env}}\cup\mathcal{T}_{\mathrm{ctx}}
       \cup\{\operatorname{answer}\} .
\end{aligned}
\end{equation}
Here $\mathcal{T}_{\mathrm{ctx}}$ contains the two dedicated workspace tools.
Recovery uses an ordinary file or terminal read from the archive path and is
therefore an environment action, which we denote abstractly by
$\operatorname{read\_path}(h,q)\in\mathcal{T}_{\mathrm{env}}$. Thus context
management is not a separate post-hoc controller. It is part of the model's
action space, conditioned on the dashboard and task evidence still visible in
$C_t$.

\methodhead{Mode switch.}
The dashboard is the proprioceptive channel of the meta-tool. From prompt text
alone, a model cannot reliably infer how costly a block is, how recent it is, or
whether it has been used again. These signals determine whether a block should
stay visible, be archived, or be recovered. In normal mode, the agent may call
environment tools, continue the task, archive blocks, or read archived payloads.
In overflow mode, ordinary tool calls are disabled until the agent reduces the
visible context. The allowed action set is therefore
\begin{equation}
\label{eq:overflow}
  \mathcal{A}_t =
  \begin{cases}
    \mathcal{T}_{\mathrm{env}}\cup\mathcal{T}_{\mathrm{ctx}}
      \cup\{\operatorname{answer}\},
      & |\widetilde C_t|\le B,\\
    \mathcal{T}_{\mathrm{ctx}},
      & |\widetilde C_t|>B .
  \end{cases}
\end{equation}
The hard budget is enforced by the harness, but the agent chooses what to move.
A final preflight guard can offload already-visible raw tool-result blocks near
the hard limit; if a newly returned result still cannot be admitted, the harness
rejects that result instead of silently truncating it. The complete loop is
given in Appendix~\ref{app:smc-loop}.

\subsection{Meta Context Tool}

\methodhead{Archive interface.}
The right panel shows the meta-context tool. Archiving takes a selected block
set $\mathcal{S}$ and a short replacement summary $\rho$:
\begin{equation}
\label{eq:archive-contract}
  \begin{aligned}
  h &= \operatorname{archive}(\mathcal{S},\rho),\\
  \operatorname{read\_path}(h,\varnothing) &\equiv \operatorname{payload}(\mathcal{S}) .
  \end{aligned}
\end{equation}
The first line creates a compact handle in the visible stream; the second states
the lossless contract using an ordinary environment read, with no third context
tool. The exact payload moves to the external archive, while $h$ keeps
the path, level, size, and checksum metadata visible, leaving a natural boundary
for external auditing~\citep{xu2026internal}.
The optional selector $q$ requests an exact range or subset of that payload;
$q=\varnothing$ requests the full payload.
Recovery is performed through ordinary file or terminal access to the stored
payload path. There is no task-specific retrieval oracle. If the payload is too
large, the agent may read bounded chunks or rerun the original source tool with
narrower arguments.

The stored payload is a transcript of what the model saw, not a guarantee that
the source was complete. If a source response was paginated or truncated, the
archive preserves that result exactly and leaves re-querying to the agent.

\methodhead{Hierarchical recovery.}
Archiving is hierarchical. A first archive level may group several raw blocks
into a bundle, as B6-9 does in the figure. Later, groups can themselves be
archived into coarser handles when context pressure grows. The visible stream
therefore stores a small summary and retrieval guide, while the hidden
trajectory stores the exact evidence. This is why \method{} differs from
summarization. Summaries guide navigation, but they are not the only
representation of the evidence.

Recovery follows the hierarchy in reverse. The agent may inspect a coarse
handle, recover the payload, and then decide whether a narrower
piece of evidence should return to active context. It need not reload a long
transcript when one row or identifier is needed. It can recover the file, search
or read a bounded part, and continue with a smaller block.

\subsection{RL Adaptation}
\label{sec:rl-method}

The interface above is training-free: an instruction-following model can use the
dashboard and context tools without changing its parameters. We additionally
consider a post-training branch that refines \emph{when} and
\emph{how} the same policy invokes archive and recovery. This branch leaves the
workspace, dashboard, and lossless tool contract unchanged. Following the
trajectory-level optimization pattern of Context-Folding~\citep{contextfolding2025},
we optimize complete interaction trajectories while conditioning every model
token on the workspace that was actually visible when that token was generated.

\methodhead{Workspace-consistent rollouts.}
For a training task $g$, the old policy samples a group of $G$ trajectories
$\{\tau_i\}_{i=1}^G$. Let $W_{i,t}$ be the workspace immediately before token
$x_{i,t}$ and let $F(W_{i,t})$ denote the rendering in
Eq.~\eqref{eq:workspace-render}. Archive actions remove their selected payloads
from subsequent renderings, whereas recovery actions restore only the content
actually read. Let $H_{i,t}:=F(W_{i,t})\oplus x_{i,<t}^{\mathrm{turn}}$ include
the autoregressive prefix already generated in the current LLM turn. The
importance ratio is therefore
\begin{equation}
\label{eq:workspace-ratio}
r_{i,t}(\theta)=
\frac{\pi_\theta(x_{i,t}\mid H_{i,t})}
     {\pi_{\mathrm{old}}(x_{i,t}\mid H_{i,t})},
\end{equation}
and $m_{i,t}=\mathbf{1}_{\mathrm{LLM}}(x_{i,t})$ masks tool observations. Thus
rollout and policy optimization use the same managed context; reconstructing the raw,
append-only transcript for the actor update would train a different policy.

\methodhead{Outcome reward and process penalties.}
Each trajectory receives a task reward $R_i\in[0,1]$ from the benchmark scorer.
Sparse task reward determines whether management preserved useful evidence. We
add only penalties for objective workspace failures; we do not reward archive
frequency or compression directly. Let
$p_{i,t}=|F(W_{i,t})|/B$ be active-context pressure and $\rho$ a soft pressure
threshold. The token-level process signal is
\begin{equation}
\label{eq:process-reward}
\begin{aligned}
Q_{i,t}={}&Q^{\mathrm{pressure}}_{i,t}
          +Q^{\mathrm{invalid}}_{i,t}
          +Q^{\mathrm{undo}}_{i,t},\\
Q^{\mathrm{pressure}}_{i,t}={}&
-\lambda_{p}\max\!\left(0,\frac{p_{i,t}-\rho}{1-\rho}\right).
\end{aligned}
\end{equation}
Pressure is computed once from the pre-action rendering---including visible
blocks, handles, notices, and dashboard---and shared by all LLM tokens in that
turn; a workspace transition affects pressure from the next turn onward.
$Q^{\mathrm{invalid}}_{i,t}=-\lambda_i$ iff the context-tool executor returns an
error or the canonical workspace hash is unchanged, covering missing handles,
malformed calls, and exact repetitions. $Q^{\mathrm{undo}}_{i,t}=-\lambda_u$
marks an otherwise valid archive--recovery pair that restores the same workspace
hash before any intervening environment observation or context growth; it is
assigned post hoc to the LLM tokens of both action turns. The three terms add
when they co-occur and are clipped only in Eq.~\eqref{eq:context-grpo-advantage}.
We use $\rho=0.8$, $\lambda_p=\lambda_i=1$, and $\lambda_u=0.2$. Archiving below
a fixed occupancy threshold is not penalized: a small prompt may still contain
low-density content worth externalizing proactively.

We form an outcome-normalized, process-shaped token advantage
\begin{equation}
\label{eq:context-grpo-advantage}
\widehat A_{i,t}=
\frac{\operatorname{clip}(R_i+Q_{i,t},0,1)-\mu_R}
     {\sigma_R+\epsilon},
\quad
\mu_R=\frac{1}{G}\sum_{j=1}^{G}R_j,
\end{equation}
where
$\sigma_R:=\sqrt{G^{-1}\sum_{j=1}^G(R_j-\mu_R)^2}$ is the within-task outcome
standard deviation. Thus the group baseline and scale come from outcome reward,
while $Q_{i,t}$ supplies local process shaping. The policy maximizes the clipped
objective
\begin{equation}
\label{eq:context-grpo-objective}
J_{\mathrm{RL}}=\mathbb E\!\left[
\frac{1}{\sum_i|\tau_i|}
\sum_{i,t}m_{i,t}\min\!\left(
r_{i,t}\widehat A_{i,t},
\operatorname{clip}(r_{i,t},1-\epsilon_{\mathrm{low}},
1+\epsilon_{\mathrm{high}})\widehat A_{i,t}
\right)\right].
\end{equation}
Here $|\tau_i|:=\sum_t m_{i,t}$ is the number of model-generated tokens in
trajectory $i$. Only model-generated tokens contribute to
Eq.~\eqref{eq:context-grpo-objective};
ordinary reasoning, context actions, recovery decisions, and final synthesis are
optimized jointly. Training prompts are screened using training-split pilot
rollouts to retain groups with both successful and unsuccessful samples, since
all-equal groups yield no outcome-relative learning signal.

This reward deliberately excludes the number of context-tool calls,
active-context reduction, and self-perception probe accuracy. Those quantities
remain held-out mechanism measurements: optimizing them directly would permit
trivial over-archiving and make any post-training perception gain circular.

\subsection{Theory: Recovery and Proprioception Are Both Necessary}
\label{sec:theory}

\method{} couples lossless recovery with an informative dashboard. The two
resources solve different problems: recovery preserves evidence after eviction,
whereas proprioception helps select what to evict. We formalize both with Fano's
inequality~\citep{sakai2020fano}: reconstruction gives the recovery separation,
and list decoding gives the information--recovery tradeoff. We keep the main
statements and intuition here; channel calculations, auxiliary lemmas, and full
proofs are in Appendices~\ref{app:theory-setup}--\ref{app:proof-sep}.

\methodhead{Recovery is necessary under budget pressure.}
In $\mathcal{T}_{N,k}$, the history contains $N$ independent $k$-bit evidence
blocks, but the prompt holds only $B<Nk$ bits. A uniformly random target block is
revealed after compression and must be reproduced exactly. A non-recovering
method retains only an in-prompt representation $R$ with $H(R)\le B$; a
recovering method may reload the selected block after the reveal.

\begin{proposition}[Recovery is necessary under budget pressure]
\label{prop:recover}
For any non-recovering method whose pre-reveal in-prompt state $R$ satisfies
$H(R)\le B$,
\[
\Pr\!\left[\text{correct on } \mathcal{T}_{N,k}\right]
\le \frac{B}{Nk} + \frac{1}{k}.
\]
\method{} is correct with probability $1$ whenever the instruction, the $N$
handles, and one recovered block fit within $B$. Thus there are feasible scaling
regimes in which the non-recovering bound vanishes while \method{} stays at $1$.
\end{proposition}

The bound captures delayed evidence demand: a lossy state cannot know which exact
block will later matter, whereas archive handles preserve every block until the
target is known.

\methodhead{Proprioception is necessary to use recovery efficiently.}
Recovery removes the losslessness barrier but not the \emph{control} problem: the
agent must still decide \emph{which} blocks to move. We give both agents the same
lossless archive and vary only how much state information identifies the best
eviction target.

\begin{definition}[Make-room instance $\mathcal{M}_{n,L,\ell}$]
\label{def:makeroom}
The working set holds $n$ blocks. A hidden index $J^\star$, uniform on
$\{1,\dots,n\}$, marks one \emph{bulky} block of size $L$; the other $n-1$
\emph{load-bearing} blocks each have size $\ell<L$, with $\kappa:=L/\ell\ge 2$ an
integer. The prompt is over budget by $L$ tokens. The bulky block is not needed
again, but every load-bearing block is queried later; after each archive the
agent observes only whether overflow remains. Sizes are net prompt tokens freed
after inserting replacement handles.
\end{definition}

To exit overflow the agent archives blocks in some order until the freed size
reaches $L$. Let $\tau$ be the position at which $J^\star$ is archived: archiving
$J^\star$ frees $L$ at once, while each load-bearing block frees only $\ell$, so
$\kappa$ of them are needed otherwise. Hence the number of load-bearing blocks
that must later be made available again is
\begin{equation}
\label{eq:Z-search}
Z \;=\; \min(\tau-1,\ \kappa).
\end{equation}
Making room efficiently is therefore a localization problem: rank $J^\star$
early and avoid unnecessary recovery obligations.

\begin{definition}[Proprioceptive interface of rate $I$]
\label{def:interface}
Before acting, the agent receives an observation $Y$ and archives in some
$(Y,U)$-measurable order, using private randomness $U\!\perp\!(J^\star,Y)$. The
\emph{rate} of the interface is the information it carries about the bulky block,
$I:=I(J^\star;Y)$ bits. A full size ledger identifies $J^\star$ and has
$I=\log_2 n$. A dashboard-free endpoint observes content-derived noisy size
signals; Proposition~\ref{prop:content} in Appendix~\ref{app:theory-setup}
bounds its induced rate. The theorem below otherwise applies to any observation
$Y$ through $I(J^\star;Y)$.
\end{definition}

\begin{theorem}[Information--recovery tradeoff]
\label{thm:sep}
Assume $2\le\kappa<n/2$ and let
$\nu:=\log_2\frac{n-\kappa}{\kappa}>0$. On
$\mathcal{M}_{n,L,\ell}$ under an interface of rate $I$
(Def.~\ref{def:interface}), the expected number $Z$ of archived load-bearing
blocks obeys
\[
\mathbb{E}[Z]\;\ge\;\kappa\left[1-\frac{I+1}{\nu}\right]_+.
\]
Here $[x]_+:=\max\{x,0\}$.
Any policy that answers every later query correctly must make all $Z$ of these
blocks available again and therefore incurs at least $\mathbb E[Z]$ expected
block-level recovery obligations. This counts blocks that must be restored, not
file-read calls: one grouped payload read may discharge several obligations at
once.
The full ledger identifies $J^\star$ and attains $Z=0$. At the other endpoint,
any low-rate interface satisfying $(I+1)/\nu\to0$ incurs
$\mathbb E[Z]/\kappa\to1$. For any fixed $\delta\in(0,1)$,
$\mathbb E[Z]\le\delta\kappa$ requires $I\ge(1-\delta)\nu-1$ bits, and the
balanced-bucket construction in Appendix~\ref{app:proof-sep} gives an
order-matching upper bound.
\end{theorem}

Thus recoverability prevents irreversible evidence loss, while state information
reduces unnecessary eviction and later restoration. The operational signature
matches the ablations: without the ledger the agent archives far more and
retrieves less ($255/57$ versus $69/105$ archive/retrieve events;
Figure~\ref{fig:ablation}b). The content-channel calculation, finite-sample
corollaries, and achievability construction are deferred to the appendix.

\section{Experiments}

\subsection{Experiment Setup}

\paragraph{Benchmarks.}
We evaluate across three online regimes with different trajectory scales:
\loca{}~\citep{locabench2026} as the primary million-token stress test,
BrowseComp-Plus~\citep{browsecompplus2025} as a 100K-scale deep-research
retrieval transfer, and GAIA~\citep{gaia2023} on a fixed 165-question validation
subset as a shorter general-assistant setting. We additionally use \ama{} as a
long-memory generalization benchmark: completed agent histories are replayed
through the \method{} workspace before question answering, testing whether the
same mechanism can operate as trajectory memory. \loca{} is external to this
work; we adopt its public 75-configuration suite and scoring protocol, counting
errors and timeouts as failures. Full benchmark protocols, subsets, scoring
rules, and budget settings are in the evaluation-details appendix.
We keep comparisons within each benchmark and do not aggregate across
heterogeneous score scales~\citep{xu2026reviewers}.

\paragraph{Baselines and configuration.}
We compare against fixed external policies, agent-mediated compression, and
production-agent baselines. The fixed-policy group includes ReAct, Tool-result
Clearing, and stale-observation masking~\citep{staleobs2026}. The
agent-mediated group includes SLIM~\citep{slim2025}, Active Context
Compression~\citep{activecontext2026}, and a structured Skeleton Compression
baseline following context-as-a-tool compressors~\citep{contextastool2025}. We
also include Context-Folding~\citep{contextfolding2025}, Auto-Archive + Recover,
and Claude Code (CLI release May 6, 2026). On \ama{}, we compare with the
benchmark AMA agent and retrieval-style memory baselines. Appendix
Table~\ref{tab:method-comparison} and the implementation appendix give the full
capability matrix, prompts, flags, dashboard format, and tool definitions.
Learned context managers whose released artifacts do not match this setting are
discussed as complementary systems and excluded from direct empirical rankings.

\method{} is training-free and uses the same strategy across backbones. On
\loca{}, the main runs use Gemini-3-Flash with a 128K budget. At each turn, the
agent sees per-block context metadata and may archive or recover exact
transcript payloads; the task tools are unchanged.
Across all \loca{} comparisons we hold fixed the task instances,
benchmark-native tools, backbone, context budget, and scoring. For
harness-native baselines, the agent loop and prompt assembly are also fixed and
only the context-management policy changes. Claude Code retains its complete
released CLI harness, with its MCP task tools replaced by the benchmark-native
equivalents. SLIM and
Active Context Compression are reproduced as training-free
inference-time baselines, while Skeleton Compression is a structured compression
baseline inspired by context-as-a-tool compressors; it is not a trained CAT
policy. \method{} uses no task-specific retrieval oracle: archived payloads are
stored as exact transcripts, and recovery is performed through ordinary file or
terminal reads from the returned archive path.

\subsection{Main Results Across Scales}

\begin{table}[t]
\centering
\begin{footnotesize}
\setlength{\tabcolsep}{3pt}
\newcommand{\mechhead}[1]{\makebox[2.65cm][c]{#1}}
\newcommand{\mechcell}[1]{\makebox[0.883cm][c]{#1}}
\newcommand{\benchhead}[1]{\makebox[2.65cm][c]{#1}}
\newcommand{\benchcell}[1]{\makebox[1.325cm][c]{#1}}
\newcommand{\grouprow}[1]{\rowcolor[HTML]{F5F5F5}\multicolumn{10}{c}{\textit{#1}}\\\midrule}
\resizebox{\textwidth}{!}{%
\begin{tabular}{@{}lccccccccc@{}}
\toprule
& \multicolumn{3}{c}{\mechhead{\textbf{Mechanism}}} & \multicolumn{2}{c}{\benchhead{\textbf{\loca{}}}} & \multicolumn{2}{c}{\benchhead{\textbf{BrowseComp-Plus}}} & \multicolumn{2}{c}{\benchhead{\textbf{GAIA}}} \\
\cmidrule(lr){2-4}\cmidrule(lr){5-6}\cmidrule(lr){7-8}\cmidrule(lr){9-10}
\textbf{Method} & \mechcell{State} & \mechcell{Ctrl} & \mechcell{Recov} & \benchcell{Acc$\uparrow$} & \benchcell{Traj$\downarrow$} & \benchcell{Acc$\uparrow$} & \benchcell{Traj$\downarrow$} & \benchcell{Acc$\uparrow$} & \benchcell{Traj$\downarrow$} \\
\midrule
\grouprow{Fixed external policy}
ReAct & \mechcell{\no} & \mechcell{\no} & \mechcell{\no} & \benchcell{22.7} & \benchcell{3.51M} & \benchcell{39.3} & \benchcell{163K} & \benchcell{61.2} & \benchcell{23K} \\
Tool-result Clearing & \mechcell{\no} & \mechcell{\no} & \mechcell{\no} & \benchcell{26.7} & \benchcell{2.60M} & \benchcell{42.7} & \benchcell{161K} & \benchcell{65.5} & \benchcell{24K} \\
Stale-obs.\ Masking & \mechcell{\no} & \mechcell{\no} & \mechcell{\no} & \benchcell{28.0} & \benchcell{3.32M} & \benchcell{38.0} & \benchcell{112K} & \benchcell{61.8} & \benchcell{24K} \\
Skeleton Compression & \mechcell{\no} & \mechcell{\no} & \mechcell{\no} & \benchcell{33.3} & \benchcell{2.84M} & \benchcell{40.0} & \benchcell{139K} & \benchcell{70.3} & \benchcell{28K} \\
\midrule
\grouprow{Agent-mediated / lossy}
SLIM (summary) & \mechcell{\no} & \mechcell{\pmark} & \mechcell{\no} & \benchcell{29.3} & \benchcell{3.76M} & \benchcell{49.3} & \benchcell{162K} & \benchcell{67.9} & \benchcell{30K} \\
Active Ctx.\ Compression & \mechcell{\no} & \mechcell{\yes} & \mechcell{\no} & \benchcell{36.0} & \benchcell{3.20M} & \benchcell{42.7} & \benchcell{162K} & \benchcell{71.5} & \benchcell{39K} \\
Context-Folding & \mechcell{\no} & \mechcell{\yes} & \mechcell{\no} & \benchcell{34.7} & \benchcell{3.41M} & \benchcell{43.3} & \benchcell{166K} & \benchcell{64.8} & \benchcell{39K} \\
\midrule
\grouprow{Lossless external store}
Auto-Archive + Recover & \mechcell{\no} & \mechcell{\no} & \mechcell{\yes} & \benchcell{44.0} & \benchcell{2.73M} & \benchcell{45.3} & \benchcell{133K} & \benchcell{63.6} & \benchcell{20K} \\
Claude Code & \mechcell{\no} & \mechcell{\yes} & \mechcell{\pmark} & \benchcell{42.7} & \benchcell{6.72M} & \benchcell{\underline{52.0}} & \benchcell{247K} & \benchcell{\textbf{73.9}} & \benchcell{44K} \\
\midrule
\grouprow{Ours and ablations}
\method{} w/o dashboard & \mechcell{\no} & \mechcell{\yes} & \mechcell{\yes} & \benchcell{37.3} & \benchcell{5.25M} & \benchcell{50.0} & \benchcell{423K} & \benchcell{68.5} & \benchcell{24K} \\
\method{} w/o recovery & \mechcell{\yes} & \mechcell{\yes} & \mechcell{\no} & \benchcell{\underline{45.3}} & \benchcell{2.99M} & \benchcell{43.3} & \benchcell{161K} & \benchcell{72.1} & \benchcell{28K} \\
\textbf{\method{} (full)} & \mechcell{\yes} & \mechcell{\yes} & \mechcell{\yes} & \benchcell{\textbf{50.7}} & \benchcell{2.86M} & \benchcell{\textbf{58.0}} & \benchcell{135K} & \benchcell{\underline{73.3}} & \benchcell{33K} \\
\bottomrule
\end{tabular}
}
\end{footnotesize}
\captionsetup{aboveskip=5pt}
\caption{\textbf{Main results across scales.} The three benchmarks span
million-, 100K-, and 10K-token trajectories, testing whether context management
transfers across operating regimes. The pattern supports our central claim:
long-horizon gains require not just compression or storage, but an agent-visible
state interface paired with controllable, lossless recovery.}
\label{tab:main-results}
\end{table}

Table~\ref{tab:main-results} compares \method{} across million-token \loca{}
trajectories, 100K-scale BrowseComp-Plus retrieval, and shorter GAIA
trajectories. On \loca{}, \method{} solves 50.7\% of tasks, versus 22.7\% for
ReAct and 42.7\% for Claude Code, while using less trajectory than Claude Code.
On BrowseComp-Plus it reaches 58.0\%, above the strongest baseline at 52.0\%.
On GAIA it remains competitive in the shorter setting, reaching 73.3\% versus
73.9\% for Claude Code. The mechanism columns separate state visibility,
decision maker, and recovery; methods missing one of these pieces do not match
the full interface consistently. The gain is largest in the settings where
evidence must survive long trajectories: \loca{} stresses repeated tool
interaction, while BrowseComp-Plus stresses retrieval followed by delayed
synthesis. The \loca{} cost ledger
also rules out a spend-more explanation: \method{} uses 2.86M tokens and 36.4
steps per task, compared with 6.72M tokens and 171.5 steps for Claude Code
(Appendix Table~\ref{tab:appendix-loca-dense}).

\begin{wraptable}[10]{r}{0.48\textwidth}
\centering
\footnotesize
\captionsetup{justification=raggedright,singlelinecheck=false}
\setlength{\tabcolsep}{2.2pt}
\renewcommand{\arraystretch}{0.92}
\begin{tabular}{@{}lcccc@{}}
\toprule
\textbf{Usage}~$\rightarrow$ & \textbf{Input} & \textbf{Output}
& \textbf{Cache} & \textbf{Cost} \\
\textbf{Method}~$\downarrow$ & \multicolumn{1}{c}{(M)}
& \multicolumn{1}{c}{(M)}
& \multicolumn{1}{c}{rate (\%)}
& \multicolumn{1}{c}{(USD)} \\
\midrule
Active Comp. & 3.158 & 0.042 & 48.3 & 1.02 \\
\method{} & 2.783 & 0.080 & 56.1 & 0.93 \\
ReAct & 3.476 & 0.031 & 36.5 & 1.26 \\
Claude Code & 7.138 & 0.156 & 92.9 & 1.05 \\
\bottomrule
\end{tabular}
\caption{\textbf{Cache-aware Gemini-3-Flash cost on \loca{}.} Input includes
cached tokens; output includes thinking.}
\label{tab:cache-cost}
\end{wraptable}

\paragraph{Cache-aware dollar cost.}
Shorter trajectories need not be cheaper if context edits invalidate the
reusable KV prefix.
We therefore combine the token ledger with standard
Gemini-3-Flash pricing: \$0.50/M uncached input, \$0.05/M cached input, and
\$3.00/M output including thinking~\citep{gemini3pricing}.
Thus \method{} remains the least expensive at \$0.93 per task while achieving
the highest accuracy, versus \$1.02 for Active Context Compression, \$1.05 for
Claude Code, and \$1.26 for ReAct.

\subsection{Pressure Regimes}

\paragraph{\loca{} million-token pressure.}

\begin{figure}[t]
\centering
\includegraphics[width=1.0\textwidth]{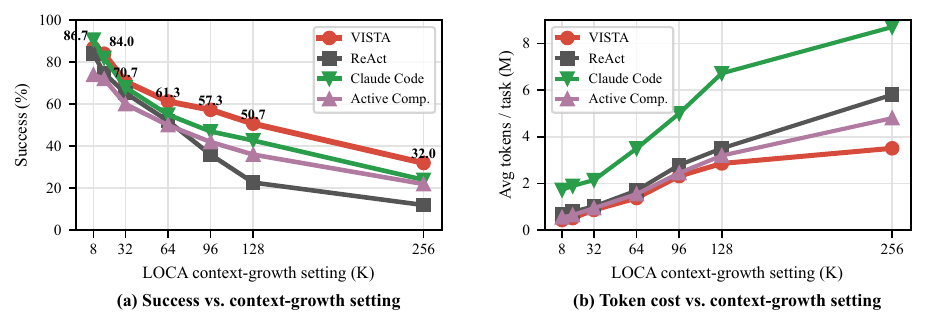}
\caption{\textbf{Pressure sweep.} Across 8K--256K context growth, \method{}
degrades more gracefully than ReAct; the right panel reports average API tokens
per task.}
\label{fig:pressure}
\end{figure}

\loca{} creates the failure mode \method{} targets: useful observations arrive
early, bulky tool results accumulate, and the agent must still act correctly many
steps later. Figure~\ref{fig:pressure} sweeps 8K to 256K context growth on the
full 75-task \loca{} suite; exact counts are in Appendix
Table~\ref{tab:loca-pressure}. The methods are close at low pressure, but the
gap opens as distractor volume grows: one-way truncation degrades while
recoverable externalization remains usable. At 8K the methods are essentially
tied (86.7 versus 84.0), but by 128K the gap is 50.7 versus 22.7, and \method{}
also spends fewer average tokens (2.86M versus 3.51M). This is the expected
signature of recoverable working memory.

\begin{figure}[t]
\centering
\begin{minipage}[t]{0.49\textwidth}
\centering
\includegraphics[width=\textwidth]{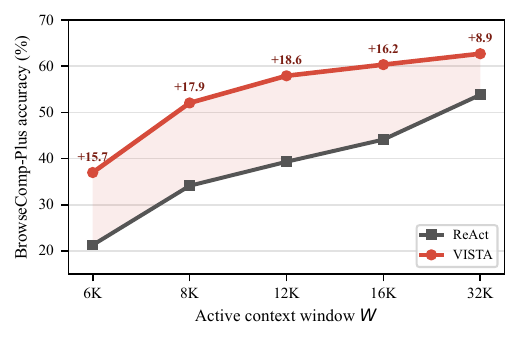}
\caption{\textbf{BrowseComp-Plus window sweep.} With task set fixed, both
methods improve as the active window grows, while \method{}'s gain over ReAct
is largest at intermediate windows.}
\label{fig:bcplus-window}
\end{minipage}\hfill
\begin{minipage}[t]{0.49\textwidth}
\centering
\includegraphics[width=\textwidth]{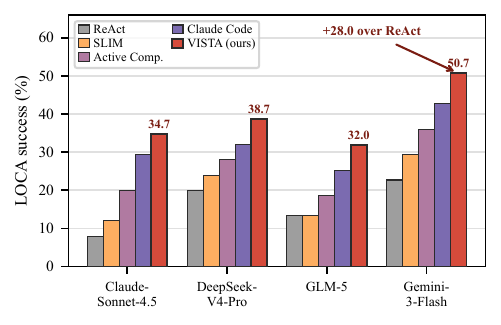}
\caption{\textbf{Cross-backbone results.} The same untrained \method{} layer is
best on all four backbones under the same benchmark task and scoring protocol.}
\label{fig:cross-model}
\end{minipage}
\end{figure}

\paragraph{BrowseComp-Plus retrieval pressure.}
\label{sec:bcplus}

BrowseComp-Plus tests whether the interface helps outside the workflow-heavy
\loca{} setting. Here the bottleneck is whether early retrieved evidence
survives until synthesis.
We set $W{=}12$K below the median first-retrieval
depth of the gold document ($\approx17$K tokens), so ReAct often loses early
evidence while \method{} continues under budget. This setting isolates
context-window pressure: loose windows usually keep the decisive document
visible, so the methods are much closer.
Figure~\ref{fig:bcplus-window} varies only $W$: tiny windows make dashboard
overhead costly, large windows let ReAct retain enough evidence, and the gain
peaks in the middle. \method{} also
uses less active context than ReAct, though more cumulative API tokens, because
it survives longer and issues more retrieval rounds.

\subsection{Backbone Robustness}

The cross-backbone result asks whether the effect is tied to one model family.
It is not: the same untrained \sms{} layer improves Claude-Sonnet-4.5,
DeepSeek-V4-Pro, GLM-5, and Gemini-3-Flash at 128K
(Figure~\ref{fig:cross-model}), including stronger backbones. This supports the
elicitation view: capable models can use a context-management interface when
runtime state is visible.

\subsection{Implicit Competence and RL Adaptation}

\begin{wraptable}[16]{r}{0.50\textwidth}
\centering
\footnotesize
\setlength{\tabcolsep}{3.2pt}
\renewcommand{\arraystretch}{0.90}
\begin{tabular}{@{}lcc@{}}
\toprule
\textbf{Method}
& \shortstack{\textbf{ID $\uparrow$}\\[-1pt]\scriptsize BC+ $\rightarrow$ BC+}
& \shortstack{\textbf{OOD $\uparrow$}\\[-1pt]\scriptsize GAIA $\rightarrow$ BC+} \\
\midrule
\rowcolor{gray!12}\multicolumn{3}{c}{\itshape Zero-shot baselines} \\
Base & 13.3 & 13.3 \\
Ours (zero-shot) & 20.0\,{\scriptsize\textcolor{visiblegreen}{(+6.7)}}
& \underline{20.0}\,{\scriptsize\textcolor{visiblegreen}{(+6.7)}} \\
\midrule
\rowcolor{gray!12}\multicolumn{3}{c}{\itshape RL adaptation} \\
Context-tool GRPO & \underline{27.3}\,{\scriptsize\textcolor{visiblegreen}{(+14.0)}}
& 17.3\,{\scriptsize\textcolor{visiblegreen}{(+4.0)}} \\
\textbf{Ours (RL)} & \textbf{31.3}\,{\scriptsize\textcolor{visiblegreen}{(+18.0)}}
& \textbf{21.3}\,{\scriptsize\textcolor{visiblegreen}{(+8.0)}} \\
\bottomrule
\end{tabular}
\caption{\textbf{Post-training and zero-shot transfer.} Our RL variant exceeds
context-tool GRPO in both regimes. Zero-shot trails ID post-training but nearly
matches OOD post-training, motivating its use in our main large-scale
experiments. Green: gain over Base.}
\label{tab:id-ood}
\end{wraptable}

Table~\ref{tab:id-ood} asks whether explicit state and a rubric for when to
manage merely replace learning, or also provide a better substrate for it. They
do both. The interface addresses the observability bottleneck, while RL can
still refine the management policy acting on the exposed state. Here
In-Distribution (ID) trains and evaluates on BrowseComp-Plus (BC+), whereas
Out-Of-Distribution (OOD) trains on GAIA and evaluates on BC+. In
distribution, the zero-shot variant improves substantially over the
Base model, while our RL variant improves further and exceeds our context-tool
GRPO baseline. For these post-training runs, we use Qwen3-8B and pre-screen
training prompts with pilot rollouts to increase the proportion of groups with
nonzero reward variance; validation and test examples are never used for this
selection. The same ordering persists when training moves to GAIA and evaluation
remains on BrowseComp-Plus: the GAIA-trained policy transfers across search
tasks, with our method retaining the best result. Thus the gain does not require
RL: zero-shot nearly matches OOD post-training before any large-scale training,
which motivates its use in our main experiments. Explicit state and
rubric-guided decisions can also be combined with GRPO, while OOD transfer shows
that the improvement is not solely training-set fit.

\begin{figure}[t]
\centering
\includegraphics[width=0.95\textwidth]{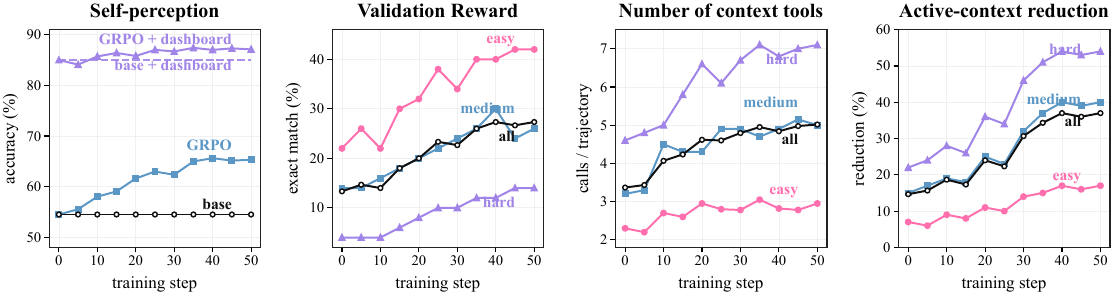}
\caption{\textbf{Context-tool GRPO learns when to manage.} Dashboard-free
perception and reward improve together, while dashboard-aided perception remains
high. Harder tasks invoke more context tools and achieve larger active-context
reductions, supporting difficulty-adaptive self-management.}
\label{fig:training-dynamics}
\end{figure}

Figure~\ref{fig:training-dynamics} localizes the implicit behavior exposed by
Figure~\ref{fig:rl-self-perception}. Hard examples use more context-tool calls
than easy ones throughout training, and tool use becomes more active as
validation reward and self-perception improve. The stratification indicates that the
learned policy does not apply one fixed management rate: it allocates more
context-management actions to harder tasks and fewer to easier ones. Their
co-movement provides mechanism evidence for difficulty-adaptive management.
Active-context reduction also rises and remains largest on hard tasks, showing
that the learned actions materially change the prompt state and do more than
merely increase search effort. Together, these curves show that RL sharpens a
state-conditioned management policy: it changes both when the agent intervenes
and how much active context it frees. The complementary zero-shot and adapted
results separate the value of exposing state from the value of optimizing the
policy that acts on it. The matched design isolates these effects under the same
backbone and evaluation protocol.

\FloatBarrier

\subsection{Offline Trajectory-Memory Transfer}

\begin{wraptable}[12]{r}{0.52\textwidth}
\centering
\begin{small}
\setlength{\tabcolsep}{3pt}
\begin{tabular}{lrrrr}
\toprule
Method & F1 & Acc. & Runtime/ep & Tokens/ep \\
\midrule
BM25       & 0.335 & 0.575 & \textbf{35.5s} & 303.43K \\
EMem & 0.363 & 0.651 & 166.2s & 470.30K \\
Mem0 & 0.329 & 0.536 & 108.0s & \textbf{30.63K} \\
AMA        & 0.368 & \textbf{0.753} & 176.5s & 268.98K \\
\method{}  & \textbf{0.382} & 0.731 & 43.7s & 148.49K \\
\bottomrule
\end{tabular}
\end{small}
\caption{\textbf{\method{} as replayed trajectory memory on AMA-Bench.} Long-memory evaluation against the specialized AMA agent and
memory-style adapters.}
\label{tab:ama-full}
\vspace*{0.8\baselineskip}
\end{wraptable}

\ama{} removes live tool interaction but preserves the long-memory demand: the
model must answer questions about completed agent histories. We adapt
\method{} by replaying each trajectory into the workspace before QA, so actions
and observations are organized into managed blocks instead of a flat prompt. This tests
whether \method{} generalizes beyond online context control to a standard
long-memory benchmark. On the full 208-episode comparison, \method{} leads on
F1, stays within about two points of the specialized AMA agent on judge accuracy,
and does so at roughly a quarter of the per-episode runtime; it also outperforms
BM25, EMem, and Mem0 adapters on F1. A training-free layer thus stays on par with
a purpose-built memory agent while running far cheaper, a clean case of transfer
to offline trajectory memory.
Table~\ref{tab:ama-domain-full} in Appendix gives the by-domain breakdown of
AMA-Bench.

\subsection{Mechanism Ablations}

\begin{figure}[t]
\centering
\includegraphics[width=0.88\textwidth]{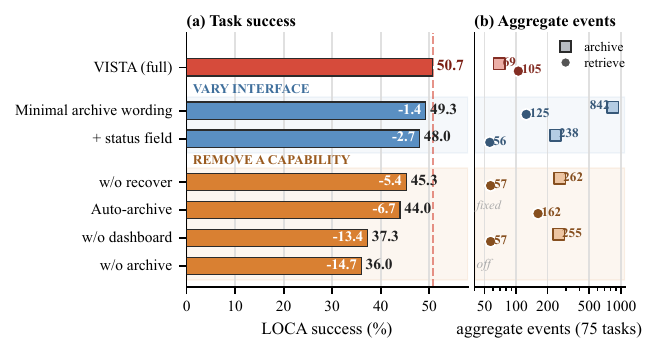}
\caption{\textbf{Component ablations.} \textbf{(a)} Removing archive, dashboard,
recovery, or agent choice hurts more than interface variants. \textbf{(b)}
Aggregate archive vs.\ retrieve events over all 75 tasks (same grouping): without the dashboard the
agent over-archives yet retrieves far less (255/57 vs.\ 69/105 for the full
system), showing blind, nonselective archiving.}
\label{fig:ablation}
\end{figure}

We probe the method along seven variants that hold the model, task set, token
estimator, prompt assembly, and context limit fixed, changing only one mechanism
at a time (Figure~\ref{fig:ablation}). Four of them remove a capability.
No-archive drops recoverable externalization, no-dashboard removes the workspace
map, no-recover hides payload paths, and fixed-archive replaces agent choice with
a static rule. Two further variants leave every capability intact and perturb
only the interface, one rephrasing the archive wording and one rendering the same
state as a status board. Because capability and description move on separate
axes, any remaining gap is attributable to the mechanism itself, not surface
wording.

The two axes come apart cleanly. Removing a capability is costly, with no-archive
falling to $27/75$ and no-dashboard to $28/75$, while the wording and
status-board variants stay within a point or two of the full $38/75$. The gain
therefore follows the capability pathway, not phrasing. Tools alone are not
enough either. No-dashboard keeps archive and recovery actions available, yet
lacking the state to target them it issues $255$ archives against only $57$
recoveries, where the full system spends $69$ archives and $105$ recoveries
(Figure~\ref{fig:ablation}b). This over-archive, under-retrieve pattern is blind
offloading, not selective management, and it is precisely the behavior the
rate-limited interface of Theorem~\ref{thm:sep} predicts.
Taken together, these interventions identify a coupled mechanism: observability
tells the agent when and what to externalize, while recoverability makes that
externalization safe. Neither interface wording nor tool availability alone
reproduces the full behavior, supporting our view of context management as
closed-loop, state-aware control.

\FloatBarrier

\section{Analysis}
\label{sec:analysis}

\paragraph{Does the perception gap actually exist?}
\begin{wraptable}[13]{r}{0.60\textwidth}
\centering
\vspace{-0.cm}
\begin{small}
\setlength{\tabcolsep}{1.5pt}
\renewcommand{\arraystretch}{0.94}
\begin{tabular}{@{}lcccccc@{}}
\toprule
& \multicolumn{2}{c}{Total size} & \multicolumn{2}{c}{Block size} & \multicolumn{2}{c}{Pairwise} \\
\cmidrule(lr){2-3}\cmidrule(lr){4-5}\cmidrule(lr){6-7}
Backbone & $-$dash & $+$dash & $-$dash & $+$dash & $-$dash & $+$dash \\
\midrule
Claude-Sonnet-4.5 & 0.84 & \textbf{0.00} & 0.37 & \textbf{0.02} & 0.67 & 0.83 \\
DeepSeek-V4-Pro   & 0.44 & \textbf{0.00} & 0.28 & \textbf{0.00} & 0.75 & 1.00 \\
GLM-5             & 0.48 & \textbf{0.00} & 0.35 & \textbf{0.00} & 0.73 & 0.88 \\
Gemini-3-Flash    & 0.43 & \textbf{0.00} & 0.24 & \textbf{0.00} & 0.68 & 1.00 \\
\bottomrule
\end{tabular}
\end{small}
\caption{\textbf{The perception gap is real and token-magnitude specific.}
Self-estimated context state with the dashboard stripped ($-$dash) versus
present ($+$dash). Size is median relative error; pairwise is within-$2\times$
accuracy.}
\label{tab:proprioception}
\vspace*{0.8\baselineskip}
\end{wraptable}

\method{} assumes an agent is not given reliable runtime metadata about its own
context. We directly test the token-magnitude component at the first archive
moment of real \loca{} runs. We
strip the dashboard and ask the backbone to report its own state along three
independent probes, namely total transcript size, individual block size, and
pairwise size comparison, scoring each against exact token counts across four
open and closed backbones (Table~\ref{tab:proprioception}). The gap is consistent
and large. Without the dashboard every backbone misjudges size, with median
relative error from $0.43$ to $0.84$ and estimates essentially uncorrelated with
truth. Adding the factual ledger reduces total-size error to zero on all four
models and block-size error to $0$--$0.02$, while lifting pairwise discrimination
toward perfect. The effect is specific to token
magnitude, not transcript memory, and it holds on open and closed weights
alike, so the intervention is a factual interface, not a stronger prompt
or a larger model. This is the empirical counterpart of the proprioceptive
channel in Theorem~\ref{thm:sep}, since the dashboard is what moves the agent
from the size-blind regime toward accurate size observability. Together, these
results establish context-state blindness in the dimension that directly
governs budgeted context management: token magnitude. VISTA exposes recency and
archive status alongside this measured signal.

\paragraph{What does a rescued run look like?}
\begin{figure}[t]
\centering
\includegraphics[width=1.0\textwidth]{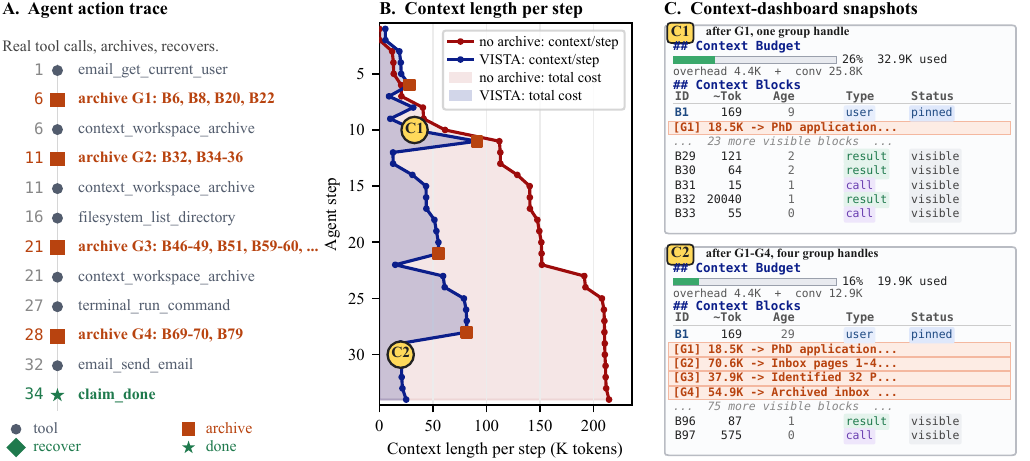}
\caption{\textbf{Case study trace.} In one 128K \loca{} run, \method{} archives
large evidence, keeps the live context compact relative to a no-archive
counterfactual, and recovers payloads when needed.}
\label{fig:case-study-curve}
\end{figure}

Figure~\ref{fig:case-study-curve} shows one 128K \loca{} email-triage run. The
dashboard marks a large inbox-export block as the biggest item, the agent
archives it, keeps the live prompt below the no-archive counterfactual, and later
reads the exact payload back for the final action. A matched baseline summarizes
or clears the block and cannot restore the verbatim value. Across rescued tasks,
this full archive-then-recover loop appears in 8 of the 16 cases where
\method{} archives at least one block. In 13 other archive-containing runs
outside the rescued subset, archiving mainly frees space without later recovery.
Thus \method{} does not win by discarding old evidence; it moves evidence out of
view while preserving an addressable recovery path. The advantage is clearest on
long trajectories, where \method{} lowers active-context overhead while improving
accuracy.

\paragraph{Does the dashboard scale?}
\begin{figure}[t]
\centering
\includegraphics[width=1.0\textwidth]{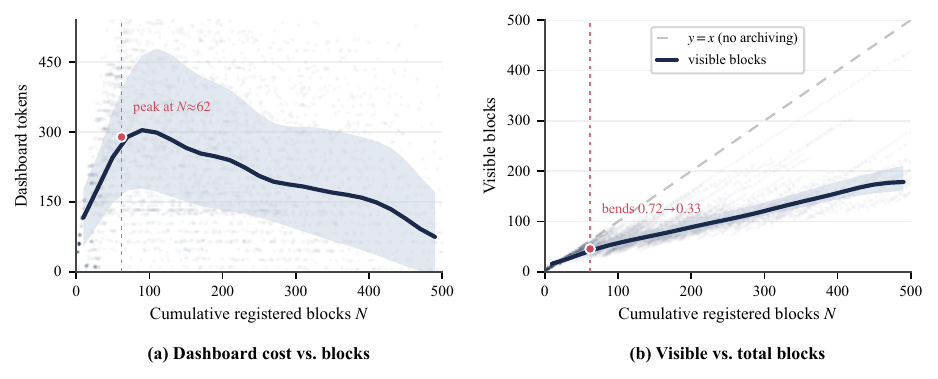}
\caption{\textbf{Dashboard overhead remains small across the observed trajectory range.}
\textbf{(a)} Net dashboard tokens (tool schemas and system prompt excluded)
against cumulative registered blocks $N$: the cost peaks near $N{\approx}62$ and
then declines over the measured range. \textbf{(b)} The visible working set
grows sublinearly in $N$ (slope bends $0.72\!\rightarrow\!0.33$ at the same
$N$). Solid lines are
binned medians; shaded bands are interquartile ranges.}
\label{fig:overhead}
\end{figure}

The dashboard adds tokens, so a fair worry is how it grows as a trajectory
registers more blocks. We measure dashboard cost on three $n{=}30$
BrowseComp-Plus runs drawn from the official $N{=}150$ evaluation
(deepseek-v4-pro, $W{=}12{,}288$, $B{=}163{,}840$). These runs are used only for
the overhead trace; the reported $58.0\%$ accuracy is computed on the full
$N{=}150$ evaluation. We use
the harness's per-turn accounting over all $30$ queries per run. Each turn logs a
fixed overhead of tool schemas plus system prompt (a constant $492$ tokens across
every query, computed with the \texttt{cl100k\_base} tokenizer) and the dashboard
message; we subtract the fixed component to isolate the dashboard's own cost.
Figure~\ref{fig:overhead}(a) shows that this net footprint is not linear in the
trajectory length: it peaks near $N{\approx}62$ at roughly $340$ tokens and then
\emph{declines} as $N$ grows over the observed range. Figure~\ref{fig:overhead}(b)
shows that the visible set also grows sublinearly, with its fitted slope bending
from $0.72$ to $0.33$ near $N{\approx}62$. Over the evaluated range, the
dashboard footprint remains small and decouples from total trajectory growth.

\section{Related Work}

Prior work clarifies where context decisions sit and what state the agent can
observe.

\paragraph{Context managed for the agent.}
One line keeps context decisions outside the model policy. Stale-observation
masking, Demand Paging, AgentOS, and AgentSwing hide, page, or route context in
the runtime
\cite{staleobs2026,demandpaging2026,agentos2026,agentswing2026}; such systems
may track size, age, or usage, but that state remains internal to the
controller. Structured-eviction and cache-efficient managers such as Context
Window Lifecycle and TokenPilot push this line further, replacing lossy
summarization with deterministic, semantically aware pruning and cache-friendly
ingestion to extend the working horizon at lower token cost
\cite{semenov2026beyond,xu2026tokenpilot}, yet they still decide what to evict on
the agent's behalf. Memory and retrieval systems such as MemGPT, Mem0, SimpleMem,
MR.Agent, PlugMem, BudgetMem, and SkillPro organize prior experience through
virtual context, long-term memory, or retrieval/graph stores
\cite{packer2023memgpt,chhikara2025mem0,simplemem2026,mragent2026,plugmem2026,budgetmem2026,skillpro2026}.
These systems provide useful storage or routing substrates, but the agent does
not receive a per-block map of its active prompt. In Figure~\ref{fig:teaser},
this corresponds to context being managed \emph{for} the agent: the runtime may
discard stale observations, page records, or retrieve memories, but the model
sees only the resulting prompt. It therefore cannot directly decide which
visible evidence should stay, move out of view, or return later, nor can it
trade a small but critical item against a large expendable tool result.

\paragraph{Self-managed context without state metadata.}
A second line makes context management part of the agent loop. Context as a Tool,
Active Context Compression, ContextBudget, Context-Folding, and LongSeeker expose
compression, folding, deletion, or routing actions, often through learned
policies or specialized controllers
\cite{contextastool2025,activecontext2026,contextbudget2026,contextfolding2025,longseeker2026}.
This line is closest to our notion of self-management: the agent or controller
can invoke context operations instead of passively waiting for truncation.
However, these operations usually act on summaries, milestones, commits, or
branches, not on a persistent block table. ContextBudget makes token
pressure explicit, and Context-Folding gives branch and return actions, but they
do not expose the VISTA-style state needed to target a precise oversized block
while preserving a small exact detail. Other systems add hierarchy, compression
guidance, agent-compatible managers, learned memory operations, or compact
reasoning summaries
\cite{hu2025hiagent,kang2025acon,yi2026learning,genericagent2026,yu2026agentic,zhang2026memory,memento2026,reasoningcache2026,markovianthinker2025,resum2025,wang2026memex},
and PACE adapts memory granularity by predicting each item's relevance to the
next action \cite{wei2026pace}.
These works support our premise that context management is an agent-level
decision, but their compressed representations are generally lossy and coarse
relative to tool traces that contain exact identifiers, URLs, or table rows.
\sms{} instead isolates what runtime state the agent must perceive, then pairs
that perception with block-level lossless archive and recovery.

\paragraph{Self-state awareness.}
Work on budget awareness, temporal blindness, and agent
externalization studies signals that models cannot infer from prompt contents
alone \cite{metacognition2026,bagen2026,temporallyblind2025,externalization2026}.
Complementary probing work finds that agent-critical information such as plans
resides in context and does not persist in hidden state, decaying once removed
from the visible context \cite{mehta2026plans}, motivating an explicit
self-description of working memory instead of assuming the model retains it.
\sms{} treats context state as such a signal and supplies it externally at
inference time. This framing separates our contribution from simply enlarging
the context window or improving summarizers: the missing information is not only
more text, but a compact self-description of the agent's current working memory.
In our setting, that self-description includes which blocks exist, how large and
old they are, whether they are visible or archived, and whether they can be
recovered exactly. This is the Figure~\ref{fig:teaser} contrast: the agent does
not merely see context, but also sees a compact state view over that context. We
evaluate this view mainly on \loca{}, which stresses online tool agents under
controllable context growth \cite{locabench2026}, and use \ama{} as a
memory-oriented transfer point \cite{ama2026}.

\section{Conclusion}

We identified context-state blindness as a basic bottleneck in long-horizon
agents: the prompt omits an explicit, reliable view of the runtime state governing
working memory. Our diagnostic establishes severe token-magnitude blindness
across four open and closed backbones, and our theory proves a sharp information
threshold for using recoverable memory efficiently.
\method{} addresses this with a runtime dashboard plus lossless archive and
recovery, giving the agent context-state information it can act on without
destroying evidence. With no training, the same interface
outperforms ReAct, deletion, masking, compaction, and Claude Code on \loca{} and
transfers across four backbones. The result shows that context management is not
only a policy-learning problem: latent capabilities can be elicited by making
hidden runtime state perceptible, positioning
the interface as a complement to post-training, not a replacement.

\bibliographystyle{plainnat}
\bibliography{references}

\clearpage
\appendix
\section{Technical Setup for the Information--Recovery Analysis}
\label{app:theory-setup}

The main text states the two separations and their operational interpretation.
Here we record the content-derived endpoint and the auxiliary list-decoding
inequality used by Theorem~\ref{thm:sep}.

For the dashboard-free endpoint of Definition~\ref{def:interface}, let the agent
observe $W_j=\ln s_j+\varepsilon_j$ for every block, with independent
$\varepsilon_j\sim\mathcal N(0,\sigma^2)$ and $\sigma>0$. This is an analyzable
model of noisy size perception, not an estimate fitted from the empirical
probe in Appendix~\ref{app:proprioception}.

\begin{proposition}[Content-derived information is bounded for fixed size ratio]
\label{prop:content}
For the channel above,
\[
I_{\mathrm{content}}=I(J^\star;W)
\le \frac{(\log_2\kappa)^2}{\sigma^2}\,\ln 2\quad\text{bits}.
\]
Thus for fixed $(\kappa,\sigma)$, content-derived size information is bounded
independently of the number of blocks $n$, whereas an exact size ledger can
identify one block among $n$.
\end{proposition}

\begin{lemma}[Fano, list-decoding form]
\label{lem:listfano}
Let $J^\star$ be uniform on $\{1,\dots,n\}$ and let
$\mathcal L(Y,U)$ be any $(Y,U)$-measurable list of size $1\le m\le n-1$,
where $U\!\perp\!(J^\star,Y)$. With
$I=I(J^\star;Y)$ and $P_c=\Pr[J^\star\in\mathcal L]$,
\[
\log_2 n-I
\le 1+P_c\log_2m+(1-P_c)\log_2(n-m).
\]
\end{lemma}

\section{Proof of Proposition~\ref{prop:recover}}
\label{app:proof}

This is Fano's inequality in \emph{reconstruction} mode: a budget-limited state
cannot reproduce more blocks than it has bits for. We restate the setting. The
history holds $N$ blocks $X_1,\dots,X_N$, each an
independent string of $k$ uniformly random bits, so $H(X_i)=k$ and the $X_i$ are
mutually independent. A non-recovering method holds a pre-reveal in-prompt state
$R$ with $H(R)\le B$. The state $R$ is a function of the blocks and the method's
internal randomness, formed before the query index $i^\star$ is revealed, and
$i^\star$ is drawn uniformly on $\{1,\dots,N\}$ independently of everything else.
After the reveal the method outputs a guess $g(R,i^\star)$, and it is correct
when $g(R,i^\star)=X_{i^\star}$.

Let $P_e^{(i)}=\Pr[g(R,i)\neq X_i]$ and let the reported success probability be
$1-P_e$ with $P_e=\frac{1}{N}\sum_{i=1}^N P_e^{(i)}$, the average over the
uniform $i^\star$.

\paragraph{Step 1: per-block Fano bound.}
Fix a block $i$. Since $X_i$ is uniform on an alphabet of size $2^k$, Fano's
inequality applied to the estimator $g(R,i)$ gives
\[
H(X_i\mid R)\;\le\; H_b\!\left(P_e^{(i)}\right) + P_e^{(i)}\log_2(2^k-1)
\;\le\; 1 + P_e^{(i)}\,k ,
\]
where $H_b$ is the binary entropy function, bounded by $1$.

\paragraph{Step 2: independence couples the blocks to a budget.}
Because the $X_i$ are mutually independent, $H(X_{1:N})=\sum_i H(X_i)$, and
subadditivity of conditional entropy gives $H(X_{1:N}\mid R)\le\sum_i H(X_i\mid
R)$. Hence
\begin{align*}
\sum_{i=1}^N I(X_i;R)
&= \sum_i\big(H(X_i)-H(X_i\mid R)\big)\\
&\le H(X_{1:N})-H(X_{1:N}\mid R)\\
&= I(X_{1:N};R)\;\le\; H(R)\;\le\; B .
\end{align*}

\paragraph{Step 3: combine.}
Using $I(X_i;R)=k-H(X_i\mid R)\ge k-1-P_e^{(i)}k$ from Step 1 and summing,
\begin{align*}
B
&\ge \sum_{i=1}^N I(X_i;R)\\
&\ge \sum_{i=1}^N\big(k-1-P_e^{(i)}k\big)
 = Nk - N - k\sum_{i=1}^N P_e^{(i)} .
\end{align*}
Dividing by $Nk$ and using $P_e=\frac1N\sum_i P_e^{(i)}$,
\begin{align*}
P_e &\ge 1 - \frac{1}{k} - \frac{B}{Nk},\\
\Pr[\text{correct}] = 1-P_e
&\le \frac{B}{Nk} + \frac{1}{k}.
\end{align*}

\paragraph{\method{} attains probability one.}
\method{} writes each block to external storage as an exact transcript and keeps
only a compact handle in the prompt. The pre-reveal prompt holds the instruction
and $N$ handles, whose size is $O(N\log N)$, not $O(Nk)$. Once $i^\star$ is
revealed, the agent reads payload $i^\star$ and recovers $X_{i^\star}$ byte for
byte. Whenever the instruction, the handles, and one recovered block fit within
$B$, the method emits $X_{i^\star}$ exactly, so its success probability is $1$.

\paragraph{Asymptotic separation.}
The gap statement requires a regime in which \method{} \emph{stays feasible}
while the lossy bound vanishes. Feasibility needs the $N$ handles plus one
recovered block to fit, i.e.\ $B\ge c_0 N\log_2 N + k$ for the constant $c_0$ set
by the handle encoding; the lossy bound $\tfrac{B}{Nk}+\tfrac1k$ vanishes when
$Nk/B\to\infty$ and $k\to\infty$. Both hold, for example, at
$k=\lceil\sqrt N\rceil$ and $B=cN\log_2 N$ with $c>c_0$: then $B$ dominates the
handle cost, so \method{} is feasible and stays at success $1$, while
$\tfrac{B}{Nk}+\tfrac1k=\Theta\!\big(\tfrac{\log N}{\sqrt N}\big)\to 0$, so the
lossy success probability tends to $0$ and the gap tends to $1$ as $N\to\infty$.
A \emph{fixed} $B$ does not exhibit this separation, because then \method{}'s own
$O(N\log N)$ handle table eventually violates the budget. The separation is thus
a statement about the growth rate of the raw evidence $Nk$ relative to a budget
$B$ that grows only fast enough to index it. \qed

\section{Proof of Proposition~\ref{prop:content}}
\label{app:proof-content}

We locate the dashboard-free endpoint of Def.~\ref{def:interface} for
$\sigma>0$. Conditioned on
$J^\star=j$, the observation $W=(W_1,\dots,W_n)$ is Gaussian with independent
coordinates, $W_i\sim\mathcal N(m_i^{(j)},\sigma^2)$, where the mean vector
$m^{(j)}$ has $m_j^{(j)}=\ln L$ (the bulky block) and $m_i^{(j)}=\ln\ell$ for
$i\neq j$. Let $P_j$ denote this conditional law and $\bar P=\frac1n\sum_{k}P_k$
the mixture.

\paragraph{Step 1: mutual information as mixture KL.}
Writing mutual information in bits and KL with natural logarithms, for
$J^\star$ uniform a standard identity gives
\[
I(J^\star;W)\;=\;\frac{1}{n\ln 2}\sum_{j=1}^n
\mathrm{KL}\!\left(P_j\,\|\,\bar P\right).
\]
Because $\mathrm{KL}(P\,\|\,\cdot)$ is convex in its second argument and
$\bar P=\frac1n\sum_k P_k$, Jensen gives
$\mathrm{KL}(P_j\|\bar P)\le\frac1n\sum_{k}\mathrm{KL}(P_j\|P_k)$, hence
\[
I(J^\star;W)\;\le\;\frac{1}{n^2\ln 2}\sum_{j=1}^n\sum_{k=1}^n
\mathrm{KL}\!\left(P_j\,\|\,P_k\right).
\]

\paragraph{Step 2: pairwise Gaussian KL.}
$P_j$ and $P_k$ are Gaussians with the same covariance $\sigma^2 \mathbf I$, so
$\mathrm{KL}(P_j\|P_k)=\frac{1}{2\sigma^2}\|m^{(j)}-m^{(k)}\|_2^2$. For $j\neq k$
the mean difference is nonzero only in coordinates $j$ and $k$: coordinate $j$
contributes $\ln L-\ln\ell=\ln\kappa$ and coordinate $k$ contributes
$\ln\ell-\ln L=-\ln\kappa$, so $\|m^{(j)}-m^{(k)}\|_2^2=2(\ln\kappa)^2$ and
\[
\mathrm{KL}(P_j\|P_k)=\frac{(\ln\kappa)^2}{\sigma^2}\quad(j\neq k),\qquad
\mathrm{KL}(P_j\|P_j)=0.
\]

\paragraph{Step 3: combine.}
There are $n(n-1)$ ordered pairs with $j\neq k$, so
\[
I(J^\star;W)\;\le\;\frac{1}{n^2\ln 2}\,n(n-1)\,
\frac{(\ln\kappa)^2}{\sigma^2}
\;\le\;\frac{(\ln\kappa)^2}{\sigma^2\ln 2}\ \text{bits}
\;=\;\frac{(\log_2\kappa)^2}{\sigma^2}\,\ln 2\ \text{bits},
\]
the last equality using $\ln\kappa=(\ln 2)\log_2\kappa$. The bound depends only on
the log size-ratio $\log_2\kappa$ and the perception noise $\sigma$. For fixed
$\kappa$ and $\sigma$, it is constant in the workspace size $n$, while the full
ledger carries $\log_2 n$ bits. More generally, Theorem~\ref{thm:sep} places this
content channel on the costly side of the make-room threshold whenever
$(I_{\mathrm{content}}+1)/\log_2(n/\kappa)\to0$. \qed

\section{Proof of Theorem~\ref{thm:sep}}
\label{app:proof-sep}

The proof has five parts: (A) the list-Fano lemma; (B) a reduction that turns the
make-room recovery cost into a list-decoding error, converting the information
budget $I$ into a lower bound on $\mathbb E[Z]$; (C) the closed-form tradeoff and
its corollaries (endpoints, threshold, price); (D) an order-wise achievability
construction; and (E) the reduction from adaptive archive policies to ordered
probing. Throughout, $J^\star$ is uniform on $[n]:=\{1,\dots,n\}$,
$\kappa=L/\ell$ is an integer satisfying $2\le\kappa<n/2$, and
$\nu:=\log_2\frac{n-\kappa}{\kappa}>0$.

\paragraph{(A) List-Fano lemma (proof of Lemma~\ref{lem:listfano}).}
Let $\mathcal L=\mathcal L(Y,U)$ be a list of size $m$ and
$E:=\mathbf 1[J^\star\notin\mathcal L]$, so $\Pr[E{=}0]=P_c$. Expand
$H(J^\star,E\mid Y,U)$ two ways. Since $E$ is a function of $(J^\star,Y,U)$,
$H(J^\star,E\mid Y,U)=H(J^\star\mid Y,U)$. Also
\[
H(J^\star,E\mid Y,U)=H(E\mid Y,U)+H(J^\star\mid E,Y,U)\le 1 + H(J^\star\mid E,Y,U).
\]
Condition on $E$: given $E{=}0$, $J^\star$ lies in a set of size $\le m$, so
$H(J^\star\mid E{=}0,Y,U)\le\log_2 m$; given $E{=}1$, $J^\star$ lies in the
complement of size $\le n-m$ (recall $\mathcal L$ is $(Y,U)$-measurable), so
$H(J^\star\mid E{=}1,Y,U)\le\log_2(n-m)$. Averaging with weights $P_c,1-P_c$ and
using $H(J^\star\mid Y,U)=H(J^\star)-I(J^\star;Y,U)=\log_2 n-I$ (as
$U\perp(J^\star,Y)$, $I(J^\star;Y,U)=I(J^\star;Y)=I$),
\[
\log_2 n - I \;\le\; 1 + P_c\log_2 m + (1-P_c)\log_2(n-m),
\]
which is Lemma~\ref{lem:listfano}. \qed

\paragraph{(B) Reduction: recovery cost lower-bounds via list decoding.}
By the ``still-over'' argument (part (E) below), it suffices to prove the lower
bound for the more powerful class of sequential policies. Such a policy archives
blocks in an order $\pi=(\pi_1,\pi_2,\dots)$ measurable in $(Y,U)$ and stops at
the first time the freed size reaches $L$. Because the bulky block frees $L$ by itself while
each load-bearing block frees only $\ell$, the process stops exactly at step
$T=\min(\tau,\kappa)$ where $\tau$ is the position of $J^\star$ in $\pi$: if
$\tau\le\kappa$ it captures $J^\star$ and frees $L$; if $\tau>\kappa$ it has
archived $\kappa$ load-bearing blocks, freeing $\kappa\ell=L$, before reaching
$J^\star$. The number of archived load-bearing blocks, which are exactly the ones
that must later be recovered, is \eqref{eq:Z-search}, $Z=\min(\tau-1,\kappa)$.

Fix any integer $1\le m\le\kappa$ and let $\mathcal L_m:=\{\pi_1,\dots,\pi_m\}$,
the first $m$ probed blocks; this is a size-$m$, $(Y,U)$-measurable list. The key
observation is
\begin{equation}
\label{eq:Z-list}
Z\ge m\ \Longleftrightarrow\ \tau>m\ \Longleftrightarrow\ J^\star\notin\mathcal L_m ,
\end{equation}
because $Z=\min(\tau-1,\kappa)\ge m$ iff $\tau-1\ge m$ (using $m\le\kappa$) iff
$J^\star$ is not among the first $m$ probes. Hence, writing
$P_c(m):=\Pr[J^\star\in\mathcal L_m]$, we have $\Pr[Z\ge m]=1-P_c(m)$.

\emph{Expectation via tail sum.} Since $Z\in\{0,1,\dots,\kappa\}$,
\begin{equation}
\label{eq:tailsum}
\mathbb E[Z]=\sum_{m=1}^{\kappa}\Pr[Z\ge m]
=\sum_{m=1}^{\kappa}\big(1-P_c(m)\big)=\kappa-\sum_{m=1}^{\kappa}P_c(m).
\end{equation}

\emph{One-shot list bound.} Apply Lemma~\ref{lem:listfano} to the size-$\kappa$
list $\mathcal L_\kappa$. With $m=\kappa$ and $n-\kappa>\kappa$ (so
$\nu=\log_2\frac{n-\kappa}{\kappa}>0$),
\[
\log_2 n-I\le 1+P_c(\kappa)\log_2\kappa+(1-P_c(\kappa))\log_2(n-\kappa).
\]
Rearranging, and using $\log_2 n\ge\log_2(n-\kappa)$,
\[
-(I+1)\ \le\ \big(\log_2(n-\kappa)-\log_2 n\big)+P_c(\kappa)\big(\log_2\kappa-\log_2(n-\kappa)\big)\ \le\ -\,P_c(\kappa)\,\nu,
\]
so $P_c(\kappa)\le\frac{I+1}{\nu}$. Because $\mathcal L_m\subseteq\mathcal
L_\kappa$ for $m\le\kappa$, $P_c(m)\le P_c(\kappa)\le\frac{I+1}{\nu}$, and
plugging into \eqref{eq:tailsum},
\begin{equation}
\label{eq:main-lb}
\boxed{\ \mathbb E[Z]\ \ge\ \kappa\left[1-\frac{I+1}{\nu}\right]_+.\ }
\end{equation}
Here the positive part combines the derived inequality with $Z\ge0$. This is the
bound in Theorem~\ref{thm:sep}.

\paragraph{(C) Corollaries.}
\emph{(i) Endpoints.} The size-aware ledger reveals the size vector, which
identifies the unique block of size $L$, so $I=\log_2 n$; then the agent archives
$\{J^\star\}$, freeing exactly $L$, giving $\tau=1$, $Z=0$, hence $0$ recoveries
and $0$ recall errors. The content-only agent (no ledger) has
$I=I_{\mathrm{content}}$. Whenever $(I_{\mathrm{content}}+1)/\nu\to0$,
\eqref{eq:main-lb} gives $\mathbb E[Z]/\kappa\to1$; fixed $\kappa$ and $\sigma$
with $n\to\infty$ is one such regime by Proposition~\ref{prop:content}. The pure
size-blind case $I=0$ is the special instance $\sigma\to\infty$.

\emph{(ii) Threshold.} $\mathbb E[Z]\le\delta\kappa$ forces, via
\eqref{eq:main-lb}, $\kappa(1-\frac{I+1}{\nu})\le\delta\kappa$, i.e.\
$I\ge(1-\delta)\nu-1$.

\emph{(iii) Price.} Before taking the positive part, the right-hand floor
$\kappa(1-\frac{I+1}{\nu})$ is affine in $I$ with slope $-\kappa/\nu$. Thus the
lower-bound floor changes at rate $\kappa/\nu$ per bit. In particular, requiring
$\mathbb E[Z]\le\kappa-\Delta$ forces
$I\ge\Delta\nu/\kappa-1$ bits.

\emph{From $Z$ to recovery obligations and errors.} Each archived load-bearing
block is queried later. Any policy that answers all these queries correctly must
make every such block available again, so the expected number of block-level
recovery obligations is $\ge\mathbb E[Z]$. This counts blocks that must be
restored, not file-read calls; a grouped payload read may satisfy several
obligations. If at most $r$ archived load-bearing blocks can be restored, at
least $[\mathbb E[Z]-r]_+$ remain unavailable in expectation, each causing an
exact-recall failure. All three size-aware costs are $0$.

\paragraph{(D) Achievability (order-wise upper bound).}
Given a rate budget $I$, let $M:=\min\{n,\lfloor2^I\rfloor\}$ and consider a
balanced $M$-bucket ledger:
partition $[n]$ into $M$ buckets, each of size at most
$b:=\lceil n/M\rceil$, and report the bucket $Y$ containing $J^\star$. Since $Y$
is a deterministic function of $J^\star$,
$I(J^\star;Y)=H(Y)\le\log_2 M\le I$, so this is a rate-$\le I$ interface. The
agent probes blocks inside the reported bucket in uniformly random order until
the flag clears. Conditional on a bucket of size $b'\le b$, the bulky block sits
at a uniform position $P\in\{1,\dots,b'\}$. As in part (E), it frees $L$ upon
capture, so the number of load-bearing blocks archived is
$\min(P-1,\kappa)\le P-1$, giving
\[
\mathbb E[Z\mid Y]\ \le\ \mathbb E[P-1\mid Y]=\frac{b'-1}{2}
\ \le\ \frac{\lceil n/M\rceil-1}{2}.
\]
The last bound holds for every bucket, and therefore also holds after averaging
over $Y$. Because $\lceil x\rceil-1<x$, $\mathbb E[Z]<n/(2M)$. Therefore
$I\ge\min\{\log_2 n,\log_2\lceil n/(2\delta\kappa)\rceil\}$ suffices for
$\mathbb E[Z]\le\delta\kappa$. At $I=\log_2 n$, we have $M=n$, hence singleton
buckets and $Z=0$. For every fixed $\delta\in(0,1)$, this sufficient rate and the
necessary rate $(1-\delta)\nu-1$ are both
$\Theta(\log_2(n/\kappa))$. If instead $\delta_n\to0$ with
$\log_2(1/\delta_n)=o(\nu)$, both rates are $(1+o(1))\nu$.

\paragraph{(E) Why archive-in-an-order is without loss, and adaptivity is useless.}
Two reductions were used above. First, any admissible batch archive policy can be
refined into a sequential order that examines the same selected blocks and stops
as soon as overflow clears. This refinement has no larger load-bearing cost, so a
lower bound proved for the more powerful sequential class also applies to batch
policies. Second, adaptivity through the binary flag adds no usable information:
before the bulky block is archived, every archived prefix of $t\le\kappa-1$
load-bearing blocks has freed $t\ell<L$, so the flag reads ``over''
deterministically and is independent of \emph{which} blocks were chosen; after
the bulky block is archived the freed size is $\ge L$ and the process stops. Thus
the only $J^\star$-information available to the ordering is $Y$, exactly as
assumed, and Lemma~\ref{lem:listfano} applies to the $(Y,U)$-measurable prefix
lists $\mathcal L_m$. This is why the lower bound \eqref{eq:main-lb} holds for all
adaptive content-only ($I=I_{\mathrm{content}}$) and rate-$I$ policies alike. \qed

\begin{remark}[Interpretation: dashboard bits versus block-level recovery]
Theorem~\ref{thm:sep} identifies a sharp, order-wise phase transition in an
information parameter: reducing the recovery burden below a fixed fraction of $\kappa$
requires $\Theta(\log(n/\kappa))$ bits about which block to evict. The affine
lower-bound floor changes at rate $\kappa/\nu$ per bit.
The no-dashboard agent perceives a genuine, noisy size signal from content, but
Proposition~\ref{prop:content} places that signal at
$I_{\mathrm{content}}\le(\ln 2)(\log_2\kappa)^2/\sigma^2$ bits. For fixed
$\kappa$ and $\sigma$ this is constant in $n$, so it falls below the
$\log_2(n/\kappa)$ make-room threshold once the workspace is large, while the
full ledger reaches the $I{=}\log_2 n$ end. The theorem itself is
distribution-free in $Y$; Proposition~\ref{prop:content} supplies an analyzable
content-only endpoint. The empirical bridge is the corresponding operational
quantity---the model's size-\emph{ranking} ability: the pairwise size-comparison
accuracy of Appendix~\ref{app:proprioception} measures how well content localizes
the block to evict, and the dashboard drives it toward the $I{=}\log_2 n$ end
(median relative total-size error $0.43$--$0.84$ without the ledger, collapsing
to $0$ with it; Table~\ref{tab:proprioception}). The $255/57$
vs.\ $69/105$ archive/retrieve split (Figure~\ref{fig:ablation}b) is the
over-archive-under-retrieve signature the theorem predicts.
\end{remark}

\FloatBarrier
\section{Method Capability Comparison}

Table~\ref{tab:method-comparison} summarizes the evaluated baselines and the
learned-compression family against the design properties of
Section~\ref{sec:methodology}. It
isolates the two properties that distinguish
\method{}: an agent-facing context dashboard and byte-exact recovery of
externalized evidence. The marks are a coarse capability summary, not a
performance claim, and partial marks reflect mechanisms that hold the property
only in part, such as Claude Code, which keeps files on disk but still
summarizes the active context.

\begin{table*}[t]
\centering
\begin{small}
\resizebox{\textwidth}{!}{%
\begin{tabular}{p{0.26\textwidth}ccccc}
\toprule
Method & Training-free & Model-agnostic & Agent-controlled & Exact recovery & Context dashboard \\
\midrule
ReAct (append until truncation) & \yes & \yes & \no & \no & \no \\
Tool-result clearing & \yes & \yes & \no & \no & \no \\
Stale-observation masking & \yes & \yes & \no & \no & \no \\
Active Context Compression & \yes & \yes & \yes & \no & \no \\
Skeleton compression & \yes & \yes & \no & \no & \no \\
Claude Code & \yes & \pmark & \yes & \pmark & \no \\
Learned compression (CAT, RL budget) & \no & \no & \yes & \no & \no \\
Context-Folding & \no & \no & \yes & \no & \no \\
LongSeeker & \no & \no & \yes & \no & \no \\
GenericAgent & \no & \no & \yes & \no & \no \\
\method{} (ours) & \yes & \yes & \yes & \yes & \yes \\
\bottomrule
\end{tabular}
}
\end{small}
\caption{\textbf{Method capability comparison.} Capability summary against the
design properties of Section~\ref{sec:methodology}; \method{} uniquely combines agent-facing state
with exact recovery.}
\label{tab:method-comparison}
\end{table*}

\section{Implementation Details}
\label{app:impl}

This section reports the exact run configuration, the verbatim prompt the agent
receives, the dashboard format, and the context tool definitions, so the setting
can be reproduced without access to our harness.

\subsection{End-to-End Context Loop}
\label{app:smc-loop}

Algorithm~\ref{alg:smc} expands the compact loop in Section~\ref{sec:methodology}.
The recoverable store $A_t$ includes agent-archived payloads and any raw tool
result externalized by the final wire-payload guard; explicit deletion removes a
payload from this store.

\begin{algorithm}[H]
\caption{Self-managed context loop}
\label{alg:smc}
\begin{algorithmic}[1]
\REQUIRE task block $b_{\mathrm{task}}$, budget $B$, environment tools
$\mathcal{T}_{\mathrm{env}}$, context tools $\mathcal{T}_{\mathrm{ctx}}$
\ENSURE final answer and workspace trajectory $W_{1:T}$
\STATE $V_0\leftarrow\{b_{\mathrm{task}}\}$; $A_0\leftarrow\varnothing$
\WHILE{the task is unfinished}
\STATE Apply the budget guard: admit fitting results; offload eligible visible
raw results or reject an unadmittable new result
\STATE $D_t\leftarrow\mathrm{Dashboard}(V_t,A_t,B)$;
$\widetilde C_t\leftarrow\mathrm{Assemble}(V_t,A_t)\cup\{D_t\}$
\STATE $C_t\leftarrow\operatorname{preflight}(\widetilde C_t,B)$
\STATE Use $\mathcal{T}_{\mathrm{ctx}}$ only if $|\widetilde C_t|>B$; otherwise
use $\mathcal{T}_{\mathrm{env}}\cup\mathcal{T}_{\mathrm{ctx}}\cup\{\mathrm{answer}\}$
\STATE $a_t\leftarrow\mathrm{LLM}(C_t)$
\STATE If $a_t=\mathrm{read\_path}(h,q)$, read the requested exact bytes and
admit the returned content subject to the same budget guard
\STATE Else if $a_t=\mathrm{archive}(\mathcal S,\rho)$, replace $\mathcal S$ by
handles and store exact payloads in $A_t$
\STATE Else if $a_t=\mathrm{delete}(\mathcal S)$, remove $\mathcal S$ and its
stored payloads
\STATE Else if $a_t\in\mathcal{T}_{\mathrm{env}}$, execute it; if
$a_t=\mathrm{answer}$, \textbf{return} the answer
\STATE $W_{t+1}\leftarrow(V_t,A_t)$
\ENDWHILE
\end{algorithmic}
\end{algorithm}

\paragraph{Benchmark setting.}
\loca{} evaluates online tool agents under controllable context growth: the
agent must continue acting while earlier reasoning, tool calls, and observations
remain in or are externalized from the working context. We evaluate 75 task
configurations; unless otherwise stated, accuracy is solved tasks over all 75,
with errors and timeouts counted as failures. \ama{} is used as a secondary
generalization benchmark. Its episodes provide a completed trajectory and ask
questions about past events, causal relations, and state changes. This is not
the native online-control setting for \method{}; it tests whether the same
context-management layer can be adapted into trajectory memory. We use the
benchmark's two-stage memory interface. During memory construction, the
completed trajectory is replayed step by step as a growing conversation: each
action and observation becomes a workspace block, the future questions are
hidden, and the replayed agent may archive exact payloads when the workspace
budget becomes tight. During retrieval, \method{} assembles the resulting
workspace, dashboard, construction events, and recoverable archive handles for
the current question. Thus the AMA-Bench result evaluates a replayed
\method{} workspace as offline memory~\citep{xu2026contextual}, instead of simply placing the full
trajectory in the model prompt.

\paragraph{Run configuration.}
\method{} is integrated into the LOCA-Bench harness and invoked as a strategy
(\texttt{loca run -s self\_managed}) with no training and no per-model tuning.
The main results use \texttt{gemini-3-flash} at a 128K budget
(\texttt{max-context-size}~$=128{,}000$); the cross-backbone runs reuse the same
strategy unchanged on \texttt{claude-sonnet-4-5}, \texttt{deepseek-v4-pro}
(open-weight), and \texttt{glm-5} (open-weight). Two flags define the full
method: \texttt{SM\_STRICT\_LONG\_CONTEXT=1} enforces a hard budget instead of
issuing a soft warning, and \texttt{SM\_BETTER\_DASHBOARD=1} selects the factual ledger
dashboard below. The ablations toggle single flags from this base, for example
\texttt{SM\_DISABLE\_ARCHIVE}, \texttt{SM\_DISABLE\_AGENT\_ARCHIVE} (fixed
archive policy), and \texttt{SM\_ENABLE\_STATE\_BOARD} (status-board variant).
Per-task timeout is 1800 seconds and reasoning effort is medium across all
backbones.

\paragraph{Baseline definitions.}
We organize the \loca{} baselines by who makes the keep-or-drop decision. Fixed
external policies include ReAct, which appends until truncation; Tool-result
Clearing, which removes old tool-result/tool-call pairs after the prompt crosses
a threshold; and fixed stale masking, which masks old tool observations while
preserving the assistant reasoning and tool-call skeleton. Agent-mediated
baselines still reduce context irreversibly. SLIM~\citep{slim2025}, reproduced
from its public release, periodically summarizes older context once the budget is
exceeded. Active Context Compression~\citep{activecontext2026} asks the agent to
write and prune its own knowledge blocks. A structured-compression baseline
preserves a compact skeleton of prior context, following the design of
context-as-a-tool compressors~\citep{contextastool2025}. Learned members of this
family generally do not release trained checkpoints and, in many cases, do not
release code, so we reproduce the training-free methods directly and follow the
published inference-time design for the rest without introducing a trained policy. Claude Code is the Claude Code command-line agent at the CLI
release of May 6, 2026, included as a strong practical agent with mature tool-use
and context-handling heuristics. We retain its complete released CLI harness and
replace only the MCP task tools with the benchmark-native equivalents. These
baselines cover deletion, masking,
summarization, self-compression, and structured compression. None combines
agent-facing context-state metadata with exact evidence recovery. On \ama{}, the
EMem-style and Mem0-style rows are local adapters implemented for this harness
and should be read as engineering baselines, not official reproductions.

\paragraph{Baseline reproduction details.}
SLIM and Active Context Compression are faithful reproductions of training-free
published methods, run with the procedure described by their authors and
triggered at the same 128K budget used for every method. SLIM periodically
summarizes older context once the budget is exceeded, and Active Context
Compression runs the explore, write a knowledge block, then prune the raw history
loop. The structured-compression baseline is inspired by context-as-a-tool
compressors but is not a faithful reimplementation, since that method is learned
and we run no trained policy. For harness-native baselines we vary only the
context-management mechanism and hold the agent loop, tools, budget, backbone,
and scoring fixed. For Claude Code, the released CLI harness is held intact
apart from substituting the benchmark-native task tools for the MCP task tools;
task instances and scoring remain unchanged.

\paragraph{Context-management protocol.}
The agent receives the following instruction block appended to the task prompt,
together with a budget notice. It is identical across backbones.

\begin{tcolorbox}[promptbox,breakable,title=Context-management protocol (verbatim)]
\begin{lstlisting}[style=prompt]
CONTEXT MANAGEMENT PROTOCOL:
A <context_workspace_status> dashboard is shown every turn as a compact map of
context blocks. Use context tools only when clearly needed. Do not archive,
delete, or offload content solely because it is old, large, or listed in context
metadata; leave content visible when the context budget is sufficient. Large
payloads may be represented by placeholders; inspect originals only when needed.
Use ordinary file/terminal/python tools, source metadata, and any in-context
payload placeholders to inspect external evidence when details are needed. For
structured data or calculations, use the source file, source tool, or query
directly. Do not copy table, CSV, or JSON rows from the conversation into code.
\end{lstlisting}
\tcblower
\begin{lstlisting}[style=prompt,basicstyle=\ttfamily\fontsize{7.2pt}{8.0pt}\selectfont]
You need to complete the task within the following context window size:
<budget:token_budget>128000</budget:token_budget>
\end{lstlisting}
\end{tcolorbox}

\paragraph{Dashboard format.}
Each turn the harness injects a \texttt{<context\_workspace\_status>} block. It is
a budget bar followed by one ledger row per block, with columns ID, approximate
tokens, age (root-turn distance, where \texttt{0r} is newest), type, compression
level, parent, and status (\texttt{visible}, \texttt{pinned}, \texttt{archived},
or \texttt{offloaded\_placeholder}). The instance below shows the same compact
column subset used in Figures~\ref{fig:teaser} and~\ref{fig:system}; the full
renderer additionally prints the compression Level and Parent columns described
above.

\begin{tcolorbox}[promptbox,breakable,title=Dashboard instance shown to the agent]
\begin{lstlisting}[style=prompt]
## Context Budget
[############--------] 62%  (~79,400 / 128,000 tokens)
  overhead ~6,200 | conversation ~71,900 | dashboard ~1,300

## Context Blocks  (Age = root-turn distance; 0r newest)
ID     ~Tok   Age  Type               Status
------------------------------------------------
B1       120   8r  user_message       pinned
B2    18,400   7r  tool_call          archived
B3     2,150   6r  assistant_message  visible
B5    14,800   3r  tool_call          visible
B6     9,300   2r  tool_call          visible
B9     1,070   0r  assistant_message  visible
\end{lstlisting}
\end{tcolorbox}

\paragraph{Context tool definitions.}
The agent acts on the workspace with two tools. Archiving replaces a block with a
compact handle and returns the payload file path; the agent recovers byte-exact
content by reading that path with ordinary file or terminal tools, so recovery is
a normal read, with no dedicated decompressor.

\begin{tcolorbox}[promptbox,breakable,title=Context tool definitions (docstrings)]
\begin{lstlisting}[style=prompt]
context_workspace_archive(block_id: str, replacement: str = "") -> str
  Replace one or more blocks with compact indexes. The original content is
  stored externally as a payload file. Operations are block-level: only listed
  block IDs are archived. Returns the payload file path for later recovery.
    block_id:    Block IDs, ranges, or group IDs, e.g. "B3", "B3,B4",
                 "B10-B20", or "G2".
    replacement: Short index text for the archived block(s).

context_workspace_delete(block_id: str, reason: str) -> str
  Permanently remove one or more blocks. Deleted content cannot be recovered.
    block_id: Block IDs, ranges, or group IDs.
    reason:   Short reason why the content has no future task value.
\end{lstlisting}
\end{tcolorbox}

Large tool results are stored as external transcript payloads with compact
placeholders. These payloads record what a tool returned to the model, not a
complete source database; if a transcript is truncated or paginated, the agent
must query the original source tool for complete data.

\section{Evaluation Details}
\label{app:expdetails}

\makeatletter
\setlength{\@fptop}{0pt}
\setlength{\@fpsep}{8pt plus 2pt}
\setlength{\@fpbot}{0pt plus 1fil}
\setlength{\@dblfptop}{0pt}
\setlength{\@dblfpsep}{8pt plus 2pt}
\setlength{\@dblfpbot}{0pt plus 1fil}
\makeatother
\renewcommand{\floatpagefraction}{0.9}
\renewcommand{\dblfloatpagefraction}{0.9}

For \loca{}, we use the independently released public task suite and evaluation
protocol without modification. It contains 75 online tool-task configurations
with controllable context growth, where prior reasoning, tool calls, and
observations accumulate until context management becomes central. We report task
success, count errors and timeouts as incorrect, and compute average steps and
tokens over task rows present in each run log. These cost values therefore
describe observed execution cost, not cost conditioned on success.
We additionally log archive and recover/read events for \method{}
variants.

\begin{figure*}[t]
\centering
\includegraphics[width=1.0\textwidth]{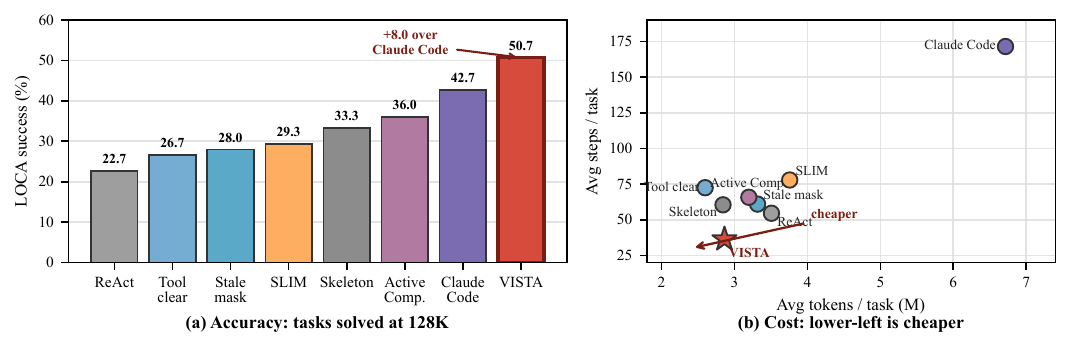}
\caption{\textbf{Expanded LOCA result at 128K.} Tasks solved, tokens, and steps
for the main \loca{} comparison.}
\label{fig:main-results}
\end{figure*}

Figure~\ref{fig:main-results} and Table~\ref{tab:appendix-loca-dense} expand
the 128K \loca{} comparison, Table~\ref{tab:rescue} gives the pairwise
rescued-task split, and Table~\ref{tab:loca-pressure} gives exact counts for the
context-growth sweep in Figure~\ref{fig:pressure}.

\begin{table*}[!tbp]
\centering
\begin{scriptsize}
\resizebox{\textwidth}{!}{%
\begin{tabular}{llrrrrrrrr}
\toprule
Family & Method & Correct & Acc. & Timeout & Error & Steps & Tokens & Mgmt. events & Notes \\
\midrule
No CM & ReAct & 17 & 22.7 & 0 & 0 & 54.6 & 3.51M & 636 trims & full 75 \\
Deletion & Tool-result Clearing & 20 & 26.7 & 0 & 0 & 72.9 & 2.60M & 1,968 clears & full 75 \\
Masking & Fixed stale masking & 21 & 28.0 & -- & -- & 61.0 & 3.32M & -- & full 75 \\
Summary & SLIM & 22 & 29.3 & -- & -- & 77.9 & 3.76M & -- & full 75 \\
Self-compression & Active Context Compression & 27 & 36.0 & -- & -- & 65.8 & 3.20M & -- & full 75 \\
Structured compression & Skeleton compression & 25 & 33.3 & -- & -- & 60.5 & 2.84M & -- & full 75 \\
Agent CLI & Claude Code & 32 & 42.7 & 0 & 22 & 171.5 & 6.72M & -- & full 75 \\
Ours & \method{} & 38 & 50.7 & 20 & 0 & 36.4 & 2.86M & 69 archive / 105 read & full 75 \\
\bottomrule
\end{tabular}
}
\end{scriptsize}
\caption{\textbf{Dense LOCA run ledger.} 128K main comparison with execution cost
and method-specific context-management events.}
\label{tab:appendix-loca-dense}
\end{table*}

\begin{table}[!tbp]
\centering
\begin{small}
\setlength{\tabcolsep}{4pt}
\begin{tabular}{lrrrr}
\toprule
Comparison & Both & Base. & \method{} & Neither \\
\midrule
Fixed stale masking & 18 & 3 & 20 & 34 \\
SLIM & 17 & 5 & 21 & 32 \\
Active Context Compression & 20 & 7 & 18 & 30 \\
Claude Code & 25 & 7 & 13 & 30 \\
\bottomrule
\end{tabular}
\end{small}
\caption{\textbf{Outcome transitions.} Pairwise 128K \loca{} split against each
baseline: both solve, baseline-only, \method{}-only, and neither.}
\label{tab:rescue}
\end{table}

\begin{table*}[!tbp]
\centering
\begin{small}
\begin{tabular}{llrrrl}
\toprule
Method & Setting & Success & Avg. steps & Avg. tok. & Notes \\
\midrule
\method{} & 8K & 86.7 & 16.7 & 0.44M & complete \\
\method{} & 16K & 84.0 & 16.2 & 0.52M & complete \\
\method{} & 32K & 70.7 & 17.4 & 0.87M & complete \\
\method{} & 64K & 61.3 & 21.2 & 1.38M & complete \\
\method{} & 96K & 57.3 & 29.7 & 2.32M & complete \\
\method{} & 128K & 50.7 & 36.4 & 2.86M & complete \\
\method{} & 256K & 32.0 & 43.1 & 3.51M & complete \\
ReAct & 8K & 84.0 & 27.9 & 0.68M & complete \\
ReAct & 16K & 74.7 & 24.9 & 0.80M & complete \\
ReAct & 32K & 65.3 & 25.3 & 1.02M & complete \\
ReAct & 64K & 52.0 & 29.6 & 1.70M & complete \\
ReAct & 96K & 36.0 & 39.5 & 2.79M & complete \\
ReAct & 128K & 22.7 & 54.6 & 3.51M & complete \\
ReAct & 256K & 12.0 & 82.6 & 5.80M & complete \\
\bottomrule
\end{tabular}
\end{small}
\caption{\textbf{Observed pressure sweep.} Matched \method{} and ReAct runs over
the full 75-task suite.}
\label{tab:loca-pressure}
\end{table*}

For BrowseComp-Plus, we evaluate deep-research retrieval with DeepSeek-V4-Pro on
an $N{=}150$ subset and report judged Pass@1 with one sampled answer. The agent
searches the benchmark's fixed corpus~\citep{browsecompplus2025}, so evidence is
scattered across retrieved passages and the transcript grows through repeated retrieval. To expose context management,
we use a deliberately tight active window ($W{=}12$K tokens per call) and total
budget ($B{=}160$K), chosen so early evidence can be evicted before synthesis.

For GAIA, we use a fixed random 165-question subset of the public validation split.
We preserve the original question text and attached files, require the official
\texttt{FINAL ANSWER:} format, and score with quasi-exact match. All methods use
DeepSeek-V4-Pro, real web/search/file tools, $W{=}12$K, and $B{=}80$K.

For \ama{}, each of the 208 episodes contains 12 open-ended questions, giving
2496 judged QA pairs. We report judge accuracy and token-level F1 following the
benchmark harness. Runtime per episode is measured for generation; judge time is
reported separately in the analysis files.

\begin{table*}[!tbp]
\centering
\begin{small}
\resizebox{\textwidth}{!}{%
\begin{tabular}{lrrrrrrrrrrrr}
\toprule
 & \multicolumn{2}{c}{Embodied} & \multicolumn{2}{c}{Game} &
\multicolumn{2}{c}{OpenQA} & \multicolumn{2}{c}{Software} &
\multicolumn{2}{c}{Text2SQL} & \multicolumn{2}{c}{Web} \\
Method & Acc. & F1 & Acc. & F1 & Acc. & F1 & Acc. & F1 & Acc. & F1 & Acc. & F1 \\
\midrule
AMA & 0.678 & 0.489 & 0.814 & 0.407 & 0.817 & 0.249 & 0.565 & 0.150 & 0.838 & 0.395 & 0.782 & 0.272 \\
\method{} & 0.683 & 0.525 & 0.747 & 0.443 & 0.731 & 0.386 & 0.567 & 0.238 & 0.846 & 0.384 & 0.763 & 0.341 \\
\bottomrule
\end{tabular}
}
\end{small}
\caption{\textbf{AMA domain breakdown.} By-domain accuracy and F1 for AMA and
the \method{} trajectory-memory adaptation.}
\label{tab:ama-domain-full}
\end{table*}

\begin{table*}[!tbp]
\centering
\begin{small}
\resizebox{\textwidth}{!}{%
\begin{tabular}{p{0.17\textwidth}p{0.19\textwidth}rp{0.18\textwidth}p{0.24\textwidth}}
\toprule
Case & Baselines failed & \method{} steps & Archive/read evidence & Evaluation signal \\
\midrule
NHL B2B schedule analysis
& ReAct, Tool-clear, SLIM
& 64
& 5 archive calls; 17 payload reads/uses
& CSV and Google Sheet correct; HA/AH/HH/AA counts verified \\
WooCommerce low-selling products
& ReAct, Claude-style, Tool-clear, SLIM
& 38
& 4 archive calls; 3 payload reads/uses
& Correct products moved; subscriber emails sent \\
Canvas final-exam schedule
& ReAct, Claude-style, Tool-clear, SLIM
& 17
& 1 archive call for course announcements
& Final Excel schedule accepted \\
NLP course reminders
& ReAct, Claude-style, Tool-clear, SLIM
& 31
& 1 archive call for large roster table
& Correct students emailed; dropped/submitted students excluded \\
\bottomrule
\end{tabular}
}
\end{small}
\caption{\textbf{Rescued-task case studies.} Tasks \method{} solves where
multiple baselines fail, with archive and payload-use counts.}
\label{tab:case-study-summary}
\end{table*}

\section{Limitations}

\sms{} supplies the missing proprioceptive signals, but it does not guarantee
the agent uses them well. A model can still misread the dashboard, archive
evidence it later needs, or recover a payload too late. The elicitation view
also predicts a floor: a model with little latent context-management skill has
little for the interface to unlock, and GLM-5 and DeepSeek-V4-Pro show the
smallest gains in our four-backbone comparison. We test four
backbones; mapping the low-capability end of this curve remains open. We also
do not test transferable adversarial inputs~\citep{xu2025one}, poisoned external
evidence~\citep{xu2025clip}, or generative trigger settings~\citep{xu2026breaking}.

\sms{} complements post-training; it is not an alternative to it.
Training improves what an agent does with context-state information, while the
dashboard supplies exact runtime state that is not explicitly available in the
prompt. Section~\ref{sec:rl-method} demonstrates this compatibility in a
Qwen3-8B transfer setting. Scaling post-training across the main multi-benchmark
and multi-backbone suite, enriching the ledger with predicted relevance, and
comparing against learned compression managers or memory-action policies
\cite{yi2026learning,yu2026agentic,zhang2026memory} are natural next steps.
Finally, the EMem-style and Mem0-style \ama{} rows are local adapters, not
official implementations, so they support diagnosis but are not final claims
against those systems; \ama{} remains a transfer test, not a primary benchmark.

Our main conclusions rest on large task-level gaps over fixed evaluation sets.
Most benchmark cells use one sampled trajectory per task; repeated-seed
evaluation remains future work for run-to-run variance and small differences.

\FloatBarrier

\section{Proprioceptive-Blindness Diagnostic}
\label{app:proprioception}

This appendix documents the diagnostic behind Table~\ref{tab:proprioception}.
The goal is to measure directly whether a backbone can read its own context
state, separating perception from skill.

\paragraph{Data.}
We anchor on the first archive event of each real \loca{} run, the moment the
agent itself decided to externalize content. We take the accumulated transcript
just before that call as the snapshot, treating each message as one block. The
runtime dashboard is not persisted in the transcript, so the stored messages are
already free of the live ledger; we additionally strip the three injected
artifacts that would leak state, namely the context-management protocol header,
the hard-limit rejection notices that print token counts, and archived-block
placeholders. A scan over all anchored snapshots confirms no residual token,
budget, or usage strings remain, and there is no per-block usage annotation.
Twenty-nine runs contain an archive; we cap each snapshot at 100K tokens by
dropping trailing blocks so it fits every backbone window, and compute ground
truth with the same tokenizer used by the harness.

\paragraph{Questions and conditions.}
\begin{wrapfigure}[22]{r}{0.45\textwidth}
\centering
\captionsetup{font=small}
\includegraphics[width=0.44\textwidth]{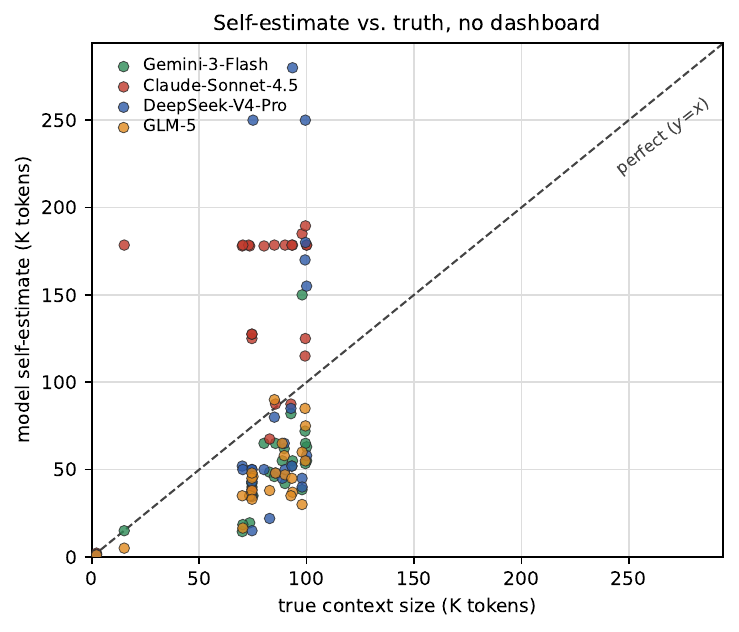}
\caption{\textbf{Proprioceptive-blindness diagnostic.} Without the dashboard,
self-estimated context size is poorly calibrated; the factual ledger closes the
gap.}
\label{fig:proprioception}
\end{wrapfigure}
We ask three quantities, each in its own request so the measurements stay
independent. \emph{Total size}: estimate the token count of the whole
transcript. \emph{Block size}: estimate the token count of four sampled blocks.
\emph{Pairwise}: for sampled block pairs, say which is larger, reported on the
hard subset within $2\times$ in true size. The
\emph{$-$dash} condition shows the cleaned transcript only; the \emph{$+$dash}
condition prepends the factual ledger (the columns of
the implementation-details appendix). Size answers are scored as median
relative error and pairwise as accuracy against the larger block. We run Gemini-3-Flash,
Claude-Sonnet-4.5, DeepSeek-V4-Pro, and GLM-5 with greedy decoding.
Claude-Sonnet-4.5 returns valid structured output slightly less often than the
other three, but the qualitative gap and its closure with the dashboard hold for
every backbone. These probes establish context-state blindness in the
token-magnitude dimension that directly governs budgeted context management;
the interface exposes recency and archive status alongside this measured signal.

\FloatBarrier

\end{document}